\documentclass[10pt]{article} 
\usepackage[preprint]{tmlr}

\usepackage{amsmath,amsfonts,bm}

\def\eqref#1{equation~\ref{#1}}

\def\1{\bm{1}}

\DeclareMathAlphabet{\mathsfit}{\encodingdefault}{\sfdefault}{m}{sl}
\SetMathAlphabet{\mathsfit}{bold}{\encodingdefault}{\sfdefault}{bx}{n}

\usepackage{hyperref}
\usepackage{url}

\usepackage{microtype}
\usepackage{graphicx}
\usepackage{subcaption}
\usepackage{booktabs} 
\usepackage{hyperref}

\usepackage{amsmath}
\usepackage{amssymb}
\usepackage{mathtools}
\usepackage{amsthm}

\usepackage{float}
\usepackage{glossaries}
\usepackage{xcolor}
\usepackage{amsmath}
\usepackage{amssymb}
\usepackage{multirow}
\usepackage{subcaption}
\usepackage{makecell}
\usepackage{lipsum} 
\usepackage{adjustbox}
\usepackage{pgfplots}
\usepgfplotslibrary{fillbetween}
\pgfplotsset{compat=1.18}

\newacronym{ttt}{TTT}{Test-Time Training}
\newacronym{ssl}{SSL}{Self-Supervised Learning}
\newacronym{gnn}{GNN}{Graph Neural Network}
\newacronym{ood}{OOD}{Out-Of-Distribution}
\newacronym{wdn}{WDN}{Water Distribution Network}
\newacronym{dt}{DT}{Digital Twin}
\newacronym{nsp}{NSP}{Next-State Prediction}
\newacronym{llm}{LLM}{Large Language Model}
\newacronym{mae}{MAE}{Mean Absolute Error}
\newacronym{mse}{MSE}{Mean Squared Error}
\newacronym{rmse}{RMSE}{Root Mean Squared Error}
\newacronym{nse}{NSE}{Nash-Sutcliffe Efficiency}
\newacronym{r2}{R$^2$}{Coefficient of Determination}
\newacronym{pcc}{PCC}{Pearson Correlation Coefficient}
\newacronym{erm}{ERM}{Empirical Risk Minimization}

\newacronym{mpnn}{MPNN}{Message Passing Neural Network}
\newacronym{mtl}{MTL}{Multi-Task Learning}
\newacronym{t3r}{T3R}{Test-Time Training with layer-wise Rotograds}
\newacronym{ogb}{OGB}{Open Graph Benchmark}
\newacronym{dwd}{DWD}{DiTEC-WDN Dataset}
\glsdisablehyper

\title{T3R: Deeper Test-Time Adaptation for
Graph Neural Networks via Gradient Rotation}

\author{\name Huy Truong \email h.c.truong@rug.nl \\
      \addr Bernoulli Institute, University of Groningen, Groningen, The Netherlands
      \AND
      \name Alexander Lazovik \email a.lazovik@rug.nl \\
      \addr Bernoulli Institute, University of Groningen, Groningen, The Netherlands
      \AND
      \name Victoria Degeler \email v.o.degeler@uva.nl\\
      \addr Informatics Institute, University of Amsterdam, Amsterdam, The Netherlands
      }

\def\month{MM}  
\def\year{YYYY} 
\def\openreview{\url{https://openreview.net/forum?id=XXXX}} 

\begin{document}

\maketitle

\begin{abstract}
\glspl{gnn} deployed in real-world systems typically have fixed weights, often leading to degraded performance under distribution shifts. This issue can be mitigated by conventional fine-tuning, but in many real-world cases, collecting labeled data is expensive or infeasible. A potential approach is \gls{ttt}, which adapts models’ weights using unlabeled test data, yet it is typically limited to shallow updates that affect only a subset of model parameters. We propose T3R, leveraging multiple Rotograd matrices to improve task affinity between the target and auxiliary tasks, essential for effective test-time training. T3R further introduces a rotation technique that reorients self-supervised signals using these matrices to create surrogate gradients for the target task, allowing deeper adaptation across nearly the entire architecture. Empirically, T3R reduces MAE by 0.172 points over standard inference in regression datasets and achieves at least 9.37\% relative improvement on cross-domain OGB classification benchmarks compared to models without adaptation. 
These results highlight the potential to develop an adaptation pipeline for graph-based systems, particularly in settings where conventional fine-tuning or retraining is infeasible.
\end{abstract}

\section{Introduction}

The recent success of deep neural networks has driven advances in model deployment. In practice, these models are typically frozen after training and used unchanged throughout their operational lifetime. However, this static deployment paradigm exposes a risk of performance degradation when encountering unseen data under distribution shifts, commonly referred to as concept drift~\citep{lu2018conceptdrift}. 
Recently, this paradigm has begun to be challenged by advances in test-time adaptation~\citep{tandon2025tttnvidia,niu2022eata,liang2021shotpp,wang2021tent}. 
Among these methods, we focus on \acrfull{ttt}, a simple yet effective approach~\citep{sun2020ttt}. It employs a Y-shaped model with a shared encoder and two heads, one for the target task and one for an auxiliary \gls{ssl}. At test time, \gls{ttt} exploits the label-free objective from the auxiliary head to adjust the model weights without access to ground-truth labels before inference. This motivates the potential of such an approach to maintain the model's capability under distribution shifts, which often occur in real-world systems.

However, this challenge is further pronounced in graph-centric, large-scale infrastructures such as 5G networks, computer networks, and water systems, where sensor drift over time can shift the data distribution and labeled samples are scarce~\citep{wang2022dtapp5g,ferriolgalmes2022dtapp2,degeler24ditec}. 
In such settings, fine-tuning is impractical. For example, in the water domain, training data are typically synthesized by simulations that require complete knowledge, an assumption rarely met in practice~\citep{truong2024gatres}. Consequently, models deployed on unseen \glspl{wdn} must adapt with limited data, insufficient for additional simulation or fine-tuning. 

Besides, since these systems are naturally graph-structured, \glspl{gnn} provide an appealing inductive bias and are therefore widely used to model them~\citep{zhang2024gnnsurrogate,kerimove2024gnnsurrogate,salem2024stateestimation}.
However, \gls{ttt} for graphs and \gls{gnn} remains in infancy. This could be due to the difficulty of choosing a graph-related \gls{ssl} task that meaningfully aligns with a specific target objective, and most \gls{ttt} variants were designed for and validated mainly on image data~\citep{sun2020ttt,liu2021tttplus,gandelsman2022tttmae}.

More critically, existing \gls{ttt} methods are fundamentally constrained, as only one component of the entire architecture, the encoder, is adapted and involved in test-time inference, while the remaining components stay fixed or do not directly participate in decision making. This restricts both efficiency and robustness under distribution shift.

In this paper, we propose \gls{t3r}, a novel adaptation approach that utilizes the idea of task-specific gradient rotation from RotoGrad~\citep{javaloy2022rotograd} at test-time. In particular, \gls{ttt}~\citep{sun2020ttt} often requires a Y-shaped deep model with a shared encoder and two heads: one for the target task and another for an auxiliary \gls{ssl} objective.
During training, \gls{t3r} simultaneously optimizes the \gls{gnn} weights and a set of learnable rotation matrices to promote alignment across two tasks. 
At test time, while typical \gls{ttt} approaches update only the encoder and \gls{ssl} decoder, \gls{t3r} additionally updates the main decoder solely using these matrices and \gls{ssl} signals, allowing a deeper adaptation over almost the entire architecture.
We evaluate \gls{t3r} on multiple regression datasets from \gls{dwd}~\citep{truong2025dwd} and classification datasets from \gls{ogb}~\citep{hu2020ogb} under \gls{ood} and cross-domain settings. The results show that \gls{t3r} consistently outperforms baseline methods on the main task.

Overall, our key contributions are as follows:
\begin{itemize}
    \item We introduce \gls{t3r} method, with a rotation technique that leverages layer-wise rotation matrices to improve main and auxiliary objectives during training, while producing surrogate gradients that enable adaptation of nearly the entire model at test time.
    
    \item We empirically select self-supervised objectives on graphs, particularly when leveraging the adaptability of \gls{ttt} to improve main task performance at test time.

    \item We conduct extensive \gls{ood} experiments across multiple regression and classification datasets, demonstrating its effectiveness, with up to 38.38\% relative improvement in RMSE on \gls{dwd} and 49.82\% in ROC-AUC on \gls{ogb}, compared to standard inference without adaptation. 

\end{itemize}

\section{Related work}
\label{sec:related_work}
\subsection{Test-Time Training}
\gls{ttt}~\citep{sun2020ttt} is a post-training approach emerging in the computer vision community to enhance the model adaptation in unseen scenarios. 
In the image domain, \cite{sun2020ttt} 
employed \textit{2-D rotation prediction}, in which the deep model was trained to predict the rotation angle of an augmented test image. Alternative \gls{ssl} tasks include Masked Autoencoders, which aim to reconstruct masked images \citep{gandelsman2022tttmae}, and contrastive learning \citep{liu2021tttplus} with an additional objective on test-time statistical properties such as mean and standard deviation estimation. Nevertheless, it remains unclear whether these methods can be applied to different modalities, such as the graph. This also raises concerns about the design of a compatible auxiliary task for the graph-related primary one.


\subsection{\acrlong{ssl} on Graphs}



The choices of auxiliary tasks vary from simple generative tasks, such as nodal attribute masking~\citep{feng2020grand}, edge masking~\citep{you2020GraphCL}, and node degree estimation~\citep{li2023maskgae}, to sophisticated tasks, including Deep Graph Infomax~\citep{velickovic2018dgi}, Graph Contrastive Coding~\citep{qiu2020gcc},  and GraphMAE~\citep{hou2022graphmae}. These advanced Graph \gls{ssl} methods often improve generalizability but come with higher costs due to increased forward passes and intensive augmentations per sample. Thus, it is crucial to choose a \gls{ssl} on graphs that can effectively balance exploration (leveraging diverse augmentations) and exploitation (allocating more adaptation steps to test-time data) within limited budgets, a challenge that our proposed method directly addresses.

\subsection{Graph Neural Networks}
\glspl{gnn} have become widely used for learning representations from graph-structured data. Early spectral approaches, such as ChebNet~\citep{defferrard2016chebnet} and GCN~\citep{kipf2017gcn}, encode graph signals into higher-dimensional representations using spectral filters. In contrast, spatial approaches such as the Message Passing Neural Network (MPNN)\citep{gilmer2017mpnn} leverage graph topology to propagate and aggregate information from local node neighborhoods. This framework has served as a foundational framework for numerous influential variants, including GAT~\citep{velickovic2018gat}, GATRes~\citep{truong2024gatres}, GIN~\citep{xu2018gin}, and DeepGCN~\citep{li2019deepgcn}. In this study, \glspl{gnn} play the role of the main architecture integrated with \gls{ttt} to tackle the distribution-shift problem. 


\subsection{Multi-Task Learning}
\gls{mtl} paradigm refers to an efficient training strategy where a single model is designed to address multiple related tasks. 
Nevertheless, jointly optimizing many objectives introduces negative knowledge transfer attributed to the conflict in gradients. One approach is to break down the joint objective into individual subproblems and devise a Pareto solution~\citep{lin2019pareto}. In addition, such a non-dominated solution could be achieved by gradient re-aggregation~\citep{navon2022nashmtl}. Alternatively, \cite{suteu2020cosreg} suggested an additional auxiliary loss term to penalize negative gradient conflicts and encourage orthogonality. Likewise, \cite{yu2020pcgrad} proposed to dissect such conflicts by adjusting the gradient directions more orthogonally. In the context of \gls{ttt}, although \gls{mtl} approaches generally do not contribute to test-time model performance, we identify RotoGrad~\citep{javaloy2022rotograd} as an exception, improving both training and test-time model performance. Furthermore, we extend its ability to generate pseudo-gradients for the main head at test time, therefore enhancing the adaptability of \glspl{gnn}.

\section{Methodology}
\label{sec:method}
We first describe essential notations for graph modality and briefly revisit \gls{ttt}. We then integrate RotoGrad, a method that encourages positive task-specific gradient alignment as a separate optimization objective during training, into \gls{ttt} by adjusting the GNN architecture and auxiliary task selection strategy. 
In light of this, we propose \gls{t3r}, which leverages gradient rotation to adapt almost the entire architecture, including the shared encoder, the \gls{ssl} decoder, and all layers of the main decoder except for the last layer, given only the input at test-time.

\begin{figure*}[!ht]
    \centering
    \includegraphics[width=\linewidth]{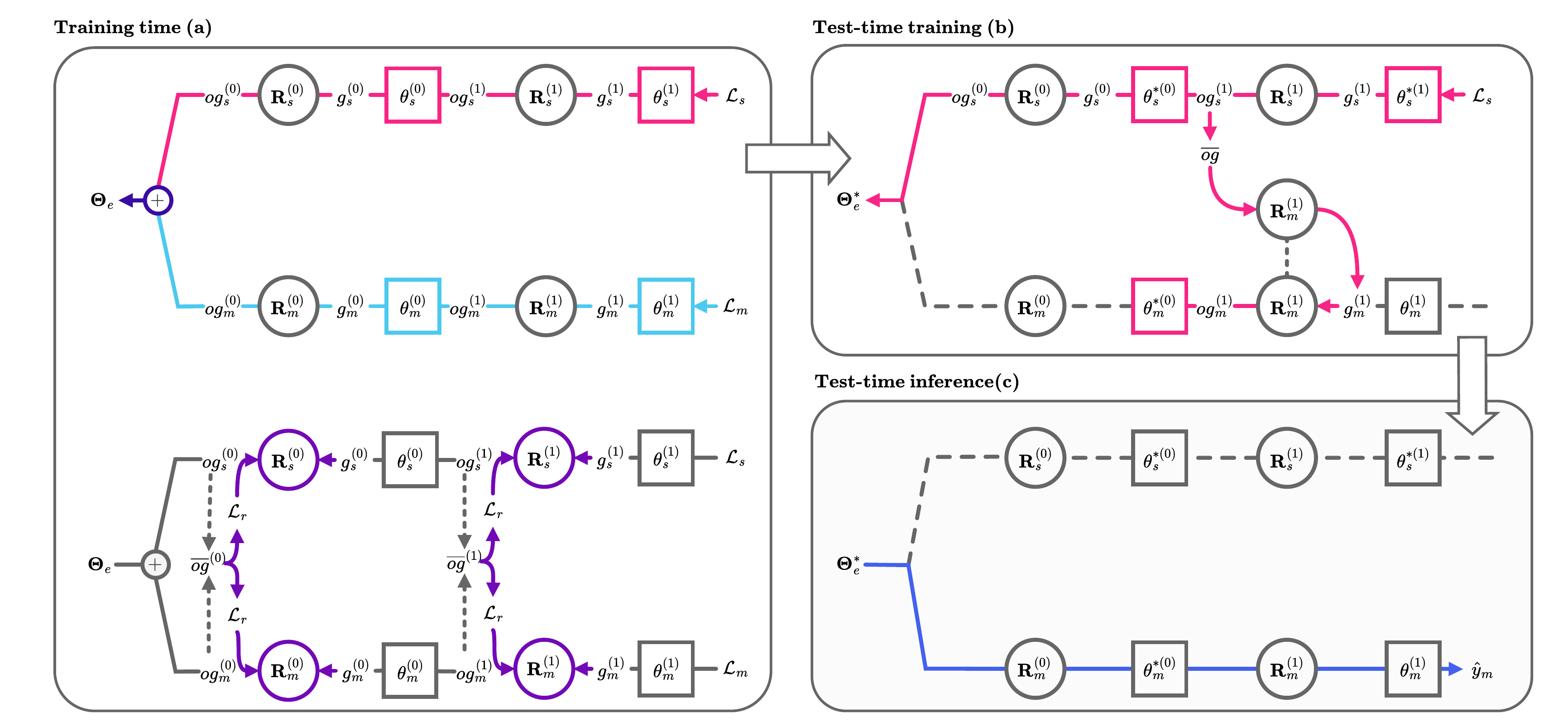}
    \caption{\textbf{Illustration of the \gls{t3r} approach using a Y-shaped architecture, where the shared encoder and both the main and \gls{ssl} decoders each consist of a two-layer \gls{gnn}.} Highlight colors indicate the gradient flow and the components updated by each loss: \textcolor[HTML]{F72585}{pink} for the auxiliary loss \(\mathcal{L}_{ssl}\), \textcolor[HTML]{4CC9F0}{cyan} for the main loss \(\mathcal{L}_{main}\), and \textcolor[HTML]{7209B7}{purple} for the rotation loss \(\mathcal{L}_{rot}\) (Equation~\ref{eqa:t3r_extension}). Subfigure (a) shows two simultaneous objectives: updating the \gls{gnn} weights (\texttt{TOP}) and the rotation matrices (\texttt{BOTTOM}). Subfigure (b) illustrates the rotation technique used to generate the surrogate gradient for the main branch. Subfigure (c) depicts the final inference involving the updated encoder and the partially updated main decoder with its fixed classifier. Leveraging rotation-aligned surrogate gradients allows a deeper adaptation of the joint model even in the absence of the main task loss.
    }
    \label{fig:t3r_vis}
\end{figure*}

\subsection{Revisiting \gls{ttt} on graphs}
We operate a joint \gls{gnn} on a graph $G=\{\mathcal{V}, \mathcal{E}\}$ in which $\mathcal{V}$ and $\mathcal{E}$ denote sets of nodes and edges, respectively. The node set is related to a nodal feature $\mathbf{X}$ while the edge set is associated with an edge feature matrix $\mathbf{E}$. The connectivity between node pairs is denoted as an adjacency matrix $\mathbf{A} \in \mathbb{R}^{|\mathcal{V}|\times |\mathcal{V}|}$, where $|\mathcal{V}|$ represents the number of nodes. 

Following \gls{ttt}, the method relies on a joint architecture consisting of three components: the encoder $\mathbf{\theta}_{e}$, the main head $\mathbf{\theta}_{m}$, and the \gls{ssl} head $\mathbf{\theta}_{s}$. These components should share the same architecture (e.g., number of layers) except for the first and last number of channels. This architecture either gains the benefits of multi-objective optimization during training or leverages self-supervised signals to partially optimize the model for unseen scenarios during inference. The primary head is designed to optimize the main task, while a secondary head targets an auxiliary objective that does not depend on raw labels and provides guidance to enhance main-task performance at test-time. Mathematically, the objective is described as:
\begin{equation} \begin{aligned} 
    \min_{\mathbf{\Theta}} \, \mathcal{L}_{main}(\mathbf{X}, \mathbf{E}, \mathbf{A}, \mathbf{Y}_{m}; \mathbf{\theta}_{e}, \mathbf{\theta}_{m}) \\
    + \lambda \mathcal{L}_{ssl}(\mathbf{X}, \mathbf{E}, \mathbf{A}, \mathbf{Y}_{s};\mathbf{\theta}_{e}, \mathbf{\theta}_{s}) 
    \end{aligned}
    \label{eqa:ttt_original}
\end{equation}
where we aim to optimize $\mathbf{\Theta} = \{\mathbf{\theta}_{e}, \mathbf{\theta}_{m}, \mathbf{\theta}_{s}\}$, and introduce the coefficient $\lambda$ to control the contribution of the auxiliary loss $\mathcal{L}_{ssl}$. This loss can be formulated in various ways. For example, in the embedding reconstruction task, the loss is calculated by representations of the original observation $\mathbf{X}$ and the augmented $\tilde{\mathbf{X}}$. In this case, the ground truth $Y_s$ is defined as $\tilde{\mathbf{X}}$. A similar computation can be applied to edge features using the ``clean'' $\mathbf{E}$ and augmented $\tilde{\mathbf{E}}$. 

During deployment, incoming data is assumed to originate from an unknown distribution that might differ from the training distribution. In a standard setting, a \gls{ttt} step can optimize $\{\theta_{s}, \theta_{e}\}$ using the \gls{ssl} signals given the test sample before the final inference:
\begin{equation} 
\label{eqa:ttt_ssl}
\mathbf{\theta}^*_{e}, \mathbf{\theta}^*_{s}=\arg\min_{\mathbf{\theta_e,\theta_s}} \, \lambda_{test} \mathcal{L}_{ssl}(\mathbf{X}, \mathbf{E}, \mathbf{A}, Y_s;\mathbf{\theta}_{e}, \mathbf{\theta}_{s}) 
\end{equation}
We refer to Equation \ref{eqa:ttt_ssl} as a single adaptation step $t$ applied to a single batch, as it can be iterated multiple times. 
The updated encoder weight $\mathbf{\theta}^*_{e}$ is then employed for the final inference by a forward pass $f$:
\begin{equation}
    \mathbf{\hat{X}} = f(\mathbf{X}, \mathbf{E}, \mathbf{A};\mathbf{\theta}^*_{e}, \mathbf{\theta}_{m})
\end{equation}

In \gls{ttt}, the main head $\mathbf{\theta}_m$ remains unchanged during testing due to the absence of supervised signals (i.e., no gradient observed). These updated weights are tied to a specific testing sample and must be reverted to the ones at the end of training time before processing the subsequent sample. 

\subsection{Task affinity measurement}
Our goal is to measure task affinity as the angle between gradient vectors of the main and \gls{ssl} heads, obtained during the training phase~\citep{du2020gradsim}.
Here, we denote the gradient $g_e, g_s, g_m$, which correspond to the encoder, the \gls{ssl} decoder, and the main decoder, respectively. Since the encoder is shared between both heads, its gradient $g_e$ is computed as a weighted sum of the other two gradients from the main and SSL tasks, implied from Equation \ref{eqa:ttt_original}. The angle $\alpha$ between two gradient vectors is calculated as:
\begin{equation} 
    \alpha = \frac{\mathbf{g}_m \cdot \mathbf{g}_s }{\|\mathbf{g}_m\| \|\mathbf{g}_s\|} = \langle \mathbf{g}_m, \mathbf{g}_s \rangle
    \label{eqa:t3r}
\end{equation}
In \gls{ssl} selection, we use cosine similarity to identify the auxiliary task that best aligns with the main task. Specifically, the auxiliary task with the highest positive gradient alignment ($\alpha$) is prioritized. Additionally, cosine similarity serves as the training objective to align task-specific gradients during training.

\subsection{Test-Time Training with layer-wise Rotograds}

Previous works have proven that positive gradient alignment $\alpha$ mitigates gradient conflict, responsible for joint-training underperformance, while improving task affinity~\citep{li2024taskaffinity}. 
In other words, a higher task affinity aligns the task gradients in the same direction, so minimizing one objective also minimizes the other.
This is essential, particularly for \gls{ttt}, as the model components are updated using \gls{ssl} signals but eventually contribute to the main task's inference at test-time.

However, in a vanilla joint architecture, such an affinity between two tasks remains fixed or slow to improve, possibly due to the gradient interference~\citep{yu2020pcgrad} and the near-orthogonality of arbitrary vectors in high-dimensional spaces~\citep{luisto2025orthogonal}.
To mitigate this, we adopt Rotograd, designed for enhancing task affinity~\citep {javaloy2022rotograd}. 
Specifically, an orthogonal rotation matrix $\mathbf{R}\in \mathcal{SO}(d)$, where $d$ is the number of hidden channels, is prepended to each decoder head in the joint model. 

We propose Test-Time Training with layer-wise Rotograds (T3R), which extends rotation matrices to every \gls{gnn} layer of two decoders in the joint model (Figure~\ref{fig:t3r_vis}). In the context of a two-task setting, we denote rotation matrices $\mathbf{R}^{(l)}_m$ and $\mathbf{R}^{(l)}_s$ for the main and \gls{ssl} heads at layer $l$, respectively. 

Considering a backward computational graph, we attempt to maximize the cosine similarity between the mean of the original (pre-rotation) gradients and task-specific  (post-rotation) gradients:
\begin{equation} \mathcal{L}_{rot}:
\max_{\mathbf{R}^{(l)}_m,\mathbf{R}^{(l)}_s} \, \langle \mathbf{R}_m^{(l)\top}\mathbf{g}^{(l)}_m,\, \overline{\mathbf{og}}^{(l)} \rangle  + \langle \mathbf{R}_s^{(l)\top} \mathbf{g}^{(l)}_s, \,\overline{\mathbf{og}}^{(l)} \rangle
\label{eqa:t3r_extension}
\end{equation}
where $\langle \cdot , \cdot \rangle$ recalls cosine similarity, $\overline{\mathbf{og}}^{(l)} = \frac{1}{2}(\mathbf{og}_m^{(l)}+\mathbf{og}_s^{(l)})$ denotes as the mean of  the main original gradient $\mathbf{og}_m^{(l)}$ and the \gls{ssl} original gradient $\mathbf{og}_s^{(l)}$. When both main and \gls{ssl} decoders have a single layer ($l=1$), Equation \ref{eqa:t3r_extension} resembles the traditional Rotograd, which mainly aims to rotate task-specific gradients to an original homogeneous direction and, therefore, makes two task gradients align closely. It is worth noting that this rotation objective $\mathcal{L}_{rot}$ is optimized cyclically alongside the joint objective in Equation~\ref{eqa:ttt_original}. This training scheme is guaranteed to converge as long as the learning rate of the rotation optimizer $\eta_{rot}$ is lower~\citep{javaloy2022rotograd}.


To optimize the per-layer rotation objectives, one could share a single optimizer across all layers. However, in practice, we use a separate optimizer for each layer, as this empirically yields a more stable learning curve. Accordingly, we scale the rotation learning rate for layer $l$ by a factor of $\frac{1}{l}$ (i.e., $\eta_{rot}^{(l)} = \frac{\eta_{rot}}{l})$.

When multiple Rotograd layers are applied and have converged, the task-specific gradients can be rotated back toward the mean original gradient $\overline{\mathbf{og}}^{(l)}$  at any layer $l$. During test-time training when the main loss $\mathcal{L}_{main}$ is absent, $\overline{\mathbf{og}}^{(l)}$ is approximately aligned with the \gls{ssl} original gradient $\mathbf{og}^{(l)}_{s}$ ($\overline{\mathbf{og}}^{(l)} \approx \mathbf{og}^{(l)}_{s}$). This allows us to employ a trick of redirecting the gradient $\mathbf{g}^{(l)}_{s}$ to the next-layer main gradient:
\begin{equation}
    \mathbf{\tilde{g}}^{(l+1)}_{m} = \mathbf{R}_s^{(l)\top} \mathbf{g}^{(l)}_s \mathbf{R}_m^{(l)}
\end{equation}
Given the surrogate main gradient $\mathbf{\tilde{g}}^{(l+1)}_{m}$, the optimizer backpropagates to the first layer of the main decoder. Note that all learnable parameters are optimized except for these main rotation matrices $\mathbf{R}^{(l)}_m$ and the last layer (i.e., classifier). We employ these updated weights of the shared encoder and partially updated main decoder following its fixed classifier to perform the final inference.

Lastly, we observe that abrupt hyperparameter shifts during test-time training and inference can be beneficial. 
In particular, we increase the coefficient $\lambda_{test}$ and reduce the augmentation intensity, such as lowering the node masking and edge dropping rates. These shifts amplify the \gls{ssl} signals for the adaptation and mitigate failure in excessively challenging \gls{ssl} tasks, for example, when strong augmentations coincide with a small number of \gls{ttt} samples.

\section{Experiments}
We evaluate model adaptability across two domains: water and molecules. We describe the selected datasets, experimental setup, and results from key experiments: \gls{ssl} task selection, regression individual benchmarks and cross-validation for \textit{next-state prediction}, and cross-domain \textit{graph property prediction} for classification. Finally, we present ablation studies of the proposed \gls{t3r} approach.


\subsection{Collections}
\label{sec:collection}
We outline the two collections in the water and molecule domains. Detailed statistics of their selected datasets are reported in Appendix~\ref{appendix:dataset_description}.

\paragraph{DiTEC-WDN~(\gls{dwd}).}~\citep{truong2025dwd} consists of numerous water networks with scenarios simulated as steady-state snapshots, under distinct configurations. Each snapshot is represented as a static, directed graph. Node features included pressure ($m$) and demand ($m^3/s$). Edge features comprise temporal attributes, flow rate ($m^3/s$), and head loss ($m$), as well as static attributes: pipe length  ($m$), diameter ($mm$), and initial status (unitless). Thirteen consecutive snapshots were grouped into non-overlapping windows. The first 12 snapshots were used as inputs, while the last snapshot was used for prediction. Datasets are pre-split into training, validation, and test sets (60/20/20).
 
\paragraph{Open Graph Benchmark~(\gls{ogb}).}~\citep{hu2020ogb} contains a variety of graph datasets in the wild. Unlike regression data in \gls{dwd}, we evaluated baselines on molecular graphs with categorical features, primarily focusing on graph property classification. In particular, we chose $\texttt{ogbg-molbace}$ and $\texttt{ogbg-molbbbp}$ for their similarity in size, input dimensionality, and number of output channels. The former dataset involves binding affinity prediction for molecules against the BACE-1 enzyme, while the latter predicts blood-brain barrier permeability. Their splits follow an 80/10/10 ratio.

\subsection{\gls{ssl} task selection via Gradient Alignment}
\label{sec:ssl_task_selection}
\paragraph{Datasets.} Two small graph datasets, \texttt{dwd-Anytown} and \texttt{dwd-19pipe} were used for this experiment. In particular, each dataset was downsampled to 10,000 non-overlapping windows. We selected 6,000 training windows from the \texttt{dwd-Anytown} training set and tested on 2,000 windows from the \texttt{dwd-19pipe} test set.

\paragraph{Baselines.} We implemented several \gls{ssl} tasks as baselines, since \gls{ttt} has seen limited use in the graph domain. To ensure efficiency, we focused on simple generative-based \gls{ssl} tasks that require at most two forward passes. 
For instance, the \textit{node embedding MSE} task computes embeddings for the original and augmented graphs in two forward passes, then measures their difference using the \gls{mse} loss.
This constraint helps mitigate the computational limitations typically encountered at inference time. These \gls{ssl} baselines are further described in the Appendix~\ref{appendix:ssl_task_description}.

\paragraph{Experiment setup.} We chose GAT~\citep{velickovic2018gat} as a backbone for all components in the joint model, using the configuration shown in Table~\ref{tab:hyper_parameters}. The only exceptions were final classifiers in both decoders, which differed based on the desired output channels for each task. The main task was fixed as \textit{next-state prediction}, while various \gls{ssl} tasks were explored, resulting in changes to the final layer of the \gls{ssl} decoder. Specifically, this layer was followed by either a global mean pooling operation for graph-level \gls{ssl} tasks or a sparse scattering after a linear transformation to output edge feature predictions for edge-level \gls{ssl} tasks. We then performed a standard \gls{ttt} step (t = 1) and compared it to standard inference (t = 0) across three runs. Additionally, we reported the average training Gradient Alignment, along with \gls{mae} and relative improvement measured on the test set.

\begin{table}[H]
    \centering
    \small
    \begin{tabular}{l|cccc}
    \hline
    \textbf{\gls{ssl} Task} & \textbf{\makecell{Test\\MAE(t=0)$\downarrow$} } & \textbf{\makecell{Test\\ MAE(t=1)$\downarrow$}} & \textbf{\makecell{Relative \\ Gain(\%)$\uparrow$}}&
    \textbf{\makecell{Train\\Gradient Alignment}} 
    \\
    \hline
    \textbf{Node mask}         & 5.8103  &\textbf{ 3.9924}  & \textbf{31.29}      & 0.12001   \\
    \textbf{Node emb Hinge }   & 5.5840  & 5.2572  & 5.85       & 0.01584   \\
    \textbf{Graph emb Hinge}   & 5.3039  & 5.0863  & 4.10       & 0.01310   \\

    \textbf{Edge mask}         & 8.0307  & 8.4091  & -4.71      & 0.00793   \\
    \textbf{Graph emb MSE}     & 5.0118  & 4.9935  & 0.37       & 0.00029   \\
    \textbf{Edge spd}          & \textbf{4.2763}  & 4.0688  & 4.85       & -0.00013  \\
    \textbf{Node emb MSE}      & 5.9647  & 5.9644  & 0.01       & -0.00033  \\
    \textbf{Node emb CLIP}     & 5.5818  & 5.5763  & 0.10       & -0.00033  \\
    \textbf{Graph emb CLIP}    & 5.6845  & 5.6555  & 0.51       & -0.00033  \\
    \textbf{Edge emb CLIP}     & 5.8712  & 5.8727  & -0.03      & -0.00033  \\
    \textbf{Edge emb MSE}      & 6.5948  & 6.6065  & -0.18      & -0.00049  \\
    \textbf{Edge emb Hinge}    & 5.1011  & 5.3939  & -5.74      & -0.00894  \\
    \hline
    \end{tabular}
    
    \caption{\textbf{\gls{ssl} Task Evaluation Based on Main Task Performance and Gradient Alignment.} The standard deviation was omitted for clarity. $t=0$ refers to a standard inference, and $t=1$ denotes a single adaptation step applied. While most \gls{ssl} tasks exhibit near-zero gradient angle with the main task, \textit{node masking} stands out as the strongest positive gradient alignment and correspondingly the highest relative gain, highlighting its superior adaptability.
    }
    \label{tab:ssl_selection}
\end{table}

\paragraph{Results.} Table~\ref{tab:ssl_selection} shows the main-task performance of \gls{ssl} tasks. While most of the auxiliary tasks are nearly orthogonal to the main task, likely a consequence of the curse of dimensionality, three tasks (\textit{node mask}, \textit{node emb hinge}, and \textit{graph emb hinge}) show a more consistent increase in relative gain when their positive alignments exceed a small threshold ($> 0.01$).

In addition, the top-scoring task, \textit{node masking}, aligns with Masked Autoencoders, a similar \gls{ssl} task previously used for \gls{ttt} in the vision domain~\citep{gandelsman2022tttmae}, and was used as the default auxiliary task in subsequent experiments. Alongside main-task performance metrics, we empirically observe that positive gradient alignment correlates with improved model adaptability. A formal theoretical justification of this relationship is left for future work.

\subsection{Next-State Prediction on \gls{dwd}}
\label{sec:nsp_on_dwd}
\paragraph{Datasets.}
After determining the best compatible task recalled as \textit{node masking}, we evaluated adaptation baselines on a regression task. In particular, there are two relevant experiments: per-benchmark evaluation and cross-validation. 
The former includes \texttt{dwd-ky4}, \texttt{dwd-ky6}, \texttt{dwd-ky7}, \texttt{dwd-npcl1}, \texttt{dwd-wa1}, \texttt{dwd-ctown}, \texttt{dwd-ky5}, \texttt{dwd-ky10}, and \texttt{dwd-ky13}, whereas the latter is restricted to a smaller set due to the significantly higher memory cost of training a joint multi-network model. Consequently, we construct a subset of five representative WDNs: (\texttt{dwd-anytown}, \texttt{dwd-jilin}, \texttt{dwd-ctown}, \texttt{dwd-wa1}, and \texttt{dwd-exn}), sampled to balance diversity across networks for the cross-validation experiment.

\begin{table}[H]
\centering
\small
\begin{adjustbox}{max width=\textwidth}
\begin{tabular}{l|ccccc}
\hline
\textbf{Method} & \textbf{R2$\uparrow$} & \textbf{RMSE$\downarrow$} & \textbf{PCC$\uparrow$} & \textbf{MAE$\downarrow$} & \textbf{NSE$\uparrow$} \\
\hline
\textbf{ERM} &  0.9051 $\pm$ 0.0783 & 0.5023 $\pm$ 0.1172 & 0.9497 $\pm$ 0.0419 & 0.3916 $\pm$ 0.0969 & -2.3600 $\pm$ 5.1430 \\
\textbf{Joint Training} & 0.8831 $\pm$ 0.1123 & 0.4782 $\pm$ 0.2135 & 0.9376 $\pm$ 0.0616 & 0.3621 $\pm$ 0.2051 & -1.0399 $\pm$ 3.8281 \\
\hline
\textbf{Tent} & 0.8985 $\pm$ 0.0842 & 0.5135 $\pm$ 0.1154 & 0.9459 $\pm$ 0.0452 & 0.3915 $\pm$ 0.0989 & -2.6587 $\pm$ 5.5148 \\
\textbf{TTT} & 0.8832 $\pm$ 0.1115 & 0.4806 $\pm$ 0.2145 & 0.9376 $\pm$ 0.0612 & 0.3640 $\pm$ 0.2045 & -1.0140 $\pm$ 3.8964 \\
\textbf{TTT + Rotograd} & 0.8541 $\pm$ 0.1310 & 0.5195 $\pm$ 0.2562 & 0.9136 $\pm$ 0.0867 & 0.4122 $\pm$ 0.2211 & -\textbf{0.1343} $\pm$ 5.0412 \\
\textbf{T3R (ours)} & \textbf{0.9492} $\pm$ 0.0827 & \textbf{0.3095} $\pm$ 0.2591 & \textbf{0.9721} $\pm$ 0.0510 & \textbf{0.2196} $\pm$ 0.2115 & -0.9875 $\pm$ 4.9584 \\
\hline
\end{tabular} 
\end{adjustbox}
\caption{\textbf{Average results over nine benchmark datasets}. A single adaptation step is applied for all methods except the first two (no-adaptation baselines). Values show mean $\pm$ pooled standard deviation over three runs across nine datasets. In terms of RMSE, T3R achieves a 38.38 \% relative improvement over ERM, while most other adaptation methods underperform ERM in this regression setting.
}
\label{tab:indiv_nsp}
\end{table}

\paragraph{Experiment setup.}
All experiments were conducted on a cluster equipped with an A100 40GB GPU. We used GAT as the backbone for all components in the joint model, following the same configuration as in the previous experiment. We reported \gls{mae}, \gls{rmse}, \gls{pcc},  \gls{r2}, and \gls{nse} (ranging from -inf to 1, higher is better), the latter being a widely used metric in hydrological modeling~\citep{legates1999nse}. 
For data augmentation, we applied node and edge dropout by default, with specific settings for the training and \gls{ttt} phases detailed in Table~\ref{tab:hyper_parameters}. Notably, we used a stronger augmentation rate during training and a weaker rate during \gls{ttt} phase, as this empirically led to better main-task performance.

\paragraph{Baselines.} 
All baselines share the same main task (\textit{next-state prediction}) and its compatible \gls{ssl} task (\textit{node masking}). Their training and test-time training (TTT) procedures are summarized below:

\begin{itemize}
    \item \textbf{\gls{erm}}: an I-shaped \gls{gnn} encoder-decoder architecture aims to optimize the primary objective. During inference time, \gls{erm} produces the final prediction without test-time training.  
    \item \textbf{Joint Training}: a Y-shaped architecture with two heads for the main and auxiliary objectives, optimized jointly. Like \gls{erm}, it does not perform any adaptation at test time.
    \item \textbf{Tent}~\citep{wang2021tent}: an I-shaped \gls{gnn} architecture aims to optimize the primary objective. At test-time, an entropy minimization loss is applied to optimize the batch norm statistics (mean and standard deviation) while other parameters remain fixed. 
    \item\textbf{\gls{ttt}}~\citep{sun2020ttt} is trained with a standard multi-objective optimization. At test time, only the encoder and \gls{ssl} decoder were updated via the \gls{ssl} loss. The final prediction is performed using an updated encoder and a fixed main decoder. 
    \textbf{\item\gls{ttt} + Rotograd}~\citep{javaloy2022rotograd} is identical to \gls{ttt} at test-time, but training objective additionally included the rotation loss $\mathcal{L}_{\text{rot}}$ to learn two rotation matrices prepended to decoders, which remain frozen during TTT phase.
    \item \textbf{\gls{t3r} (ours)} extends the \gls{gnn} architecture with layer-wise rotation matrices. At test-time, \gls{t3r} leverages the rotation technique to update nearly all model weights. The final prediction is produced by the updated encoder, the partially updated main decoder, and the fixed classifier.
\end{itemize}


\paragraph{Per-benchmark average result.}~\label{exp:per_benchmark_nsp} Table~\ref{tab:indiv_nsp} showcases baseline performance over nine benchmark datasets. Prepending two rotation matrices in the \text{TTT + Rotograd} setup offers no measurable improvement in the \gls{gnn}’s adaptability. By contrast, extending to layer-wise matrices and applying the rotation technique, \text{T3R} consistently outperforms all other baselines on every metric except NSE, likely due to anomalous results on the \texttt{dwd-wa1} dataset (see individual results in Appendix~\ref{appendix:per_dataset_results}). Surprisingly, \text{ERM} remains competitive on some metrics (e.g., $\text{R}^2$ and PCC), while other adaptation methods struggle in this regression setting.

\begin{table}[H]
\centering
\small
\begin{adjustbox}{max width=\textwidth}
    \begin{tabular}{ll|ccccc}
    \hline
    \textbf{Method} &\textbf{Step} & \textbf{R2$\uparrow$} & \textbf{RMSE$\downarrow$} & \textbf{PCC$\uparrow$} & \textbf{MAE$\downarrow$} & \textbf{NSE$\uparrow$} \\
    \hline
    
    \textbf{Tent} &$0^{\dagger}$ & 0.9307 $\pm$ 0.0729 & 0.3604 $\pm$ 0.1503 & 0.9637 $\pm$ 0.0391 & 0.2227 $\pm$ 0.0869 & -1.8122 $\pm$ 4.7780 \\
     &1 & 0.9224 $\pm$ 0.0762 & 0.3924 $\pm$  0.1539 & 0.9592 $\pm$ 0.0409 & 0.2422  $\pm$ 0.0958 &  -2.0129 $\pm$ 5.1665  \\ 	
     & 10 &  0.9224  $\pm$ 0.0762 & 0.3923  $\pm$  0.1540  & 0.9592 $\pm$ 0.0409  &  0.2421  $\pm$ 0.0960  &  -2.0127  $\pm$ 5.1661   \\ 	
     & 20 & 0.9222 $\pm$ 0.0763 & 0.3926 $\pm$ 0.1541 & 0.9591 $\pm$ 0.0410 & 0.2424 $\pm$ 0.0959 & -2.0134 $\pm$ 5.1650 \\		
     & 40 & 0.9223 $\pm$ 0.0762 & 0.3923 $\pm$ 0.1540 & 0.9592 $\pm$ 0.0409 & 0.2422 $\pm$ 0.0959 & -2.0139 $\pm$ 5.1663 \\
    \hline
        
    \textbf{TTT}  & $0^{\ddagger}$  & 0.9336 $\pm$ 0.0764 & 0.3321 $\pm$ 0.1579 & 0.9652 $\pm$ 0.0407 & 0.2075 $\pm$ 0.1004 & -0.6480 $\pm$ 2.4796 \\
      & 1 & 0.9340 $\pm$ 0.0763 & 0.3330 $\pm$ 0.1570 & 0.9654 $\pm$ 0.0406 & 0.2078 $\pm$ 0.1001 & -0.6484 $\pm$ 2.4803 \\
     & 10 & 0.9355 $\pm$ 0.0750 & 0.3493 $\pm$ 0.1412 & 0.9662 $\pm$ 0.0399 & 0.2196 $\pm$ 0.0905 & -0.6561 $\pm$ 2.4817 \\
     & 20 & 0.9363 $\pm$ 0.0726 & 0.3459 $\pm$ 0.1354 & 0.9667 $\pm$ 0.0385 & 0.2175 $\pm$ 0.0867 & -0.6543 $\pm$ 2.4769 \\
     & 40 & 0.9357 $\pm$ 0.0707 & 0.3425 $\pm$ 0.1314 & 0.9664 $\pm$ 0.0375 & 0.2158 $\pm$ 0.0837 & -0.6536 $\pm$ 2.4724 \\
    \hline
    
    \textbf{TTT + Rotograd} &  0  & 0.9525 $\pm$ 0.0388 & 0.2797 $\pm$ 0.1022 & 0.9756 $\pm$ 0.0200 & 0.1699 $\pm$ 0.0630 & 0.0486 $\pm$ 1.4518 \\
    & 1 & 0.9526 $\pm$ 0.0380 & 0.2812 $\pm$ 0.0981 & 0.9757 $\pm$ 0.0196 & 0.1712 $\pm$ 0.0607 & 0.0533 $\pm$ 1.4483 \\
     & 10 & 0.9599 $\pm$ 0.0282 & 0.2696 $\pm$ 0.0793 & 0.9795 $\pm$ 0.0144 & 0.1635 $\pm$ 0.0488 & 0.0469 $\pm$ 1.4710 \\
     & 20 & 0.9612 $\pm$ 0.0257 & 0.2694 $\pm$ 0.0738 & 0.9801 $\pm$ 0.0131 & 0.1633 $\pm$ 0.0443 & 0.0350 $\pm$ 1.4924 \\
     & 40 & 0.9622 $\pm$ 0.0235 & 0.2742 $\pm$ 0.0719 & 0.9807 $\pm$ 0.0120 & 0.1674 $\pm$ 0.0441 & 0.0119 $\pm$ 1.5374 \\
    
    \hline
    \textbf{T3R (ours)} & 0 & 0.9673 $\pm$ 0.0254 & 0.2467 $\pm$ 0.0686 & 0.9833 $\pm$ 0.0130 & 0.1523 $\pm$ 0.0416 & 0.0071 $\pm$ 1.8002 \\
     & 1 & 0.9672 $\pm$ 0.0259 & \textbf{0.2373} $\pm$ 0.0715 & 0.9832 $\pm$ 0.0133 & \textbf{0.1446} $\pm$ 0.0426 & \textbf{0.0158} $\pm$ 1.7817 \\
     & 10 & 0.9687 $\pm$ 0.0245 & 0.2410 $\pm$ 0.0740 & 0.9840 $\pm$ 0.0125 & 0.1491 $\pm$ 0.0473 & 0.0106 $\pm$ 1.6943 \\
     & 20 & 0.9689 $\pm$ 0.0251 & 0.2522 $\pm$ 0.0814 & 0.9841 $\pm$ 0.0129 & 0.1604 $\pm$ 0.0564 & -0.1068 $\pm$ 1.7659 \\
     & 40 &\textbf{0.9696} $\pm$ 0.0243 & 0.2969 $\pm$ 0.0939 & \textbf{0.9845} $\pm$ 0.0125 & 0.1959 $\pm$ 0.0649 & -0.8078 $\pm$ 3.6360 \\
    
    \hline
\multicolumn{5}{l}{$0^{\dagger}$ and $0^{\ddagger}$ indicate \gls{erm} and Joint Training without adaptation, respectively.}
    \end{tabular} 
\end{adjustbox}
\caption{\textbf{Cross-validation results with more adaptation steps.} Although T3R consistently outperforms the baselines, we observe a common trend across all methods: increasing the number of adaptation steps can lead to degraded performance in this regression task.
}
\label{tab:cv_nsp}
\end{table}

\paragraph{Cross-validation result.} In the per-benchmark experiment, the \gls{gnn} equipped with the proposed adaptation methods still encounters test-set topologies that are similar to those seen during training. To assess robustness more rigorously, we further conduct a five-fold cross-validation study. Specifically, we trained baselines on the training sets of four \glspl{wdn} and evaluated them on the test set of a held-out network. 

Table~\ref{tab:cv_nsp} highlights the performance of baselines on unseen topologies with varied adaptation steps. With a single step ($t=1$), T3R achieves its best performance, particularly in RMSE and MAE. However, deeper adaptation tends to reduce performance relative to a shallow adaptation across all methods, except for $R^2$ and PCC, which consistently improve for T3R.

\subsection{Graph Property Prediction on \gls{ogb}} 
\label{sec:gpp_on_ogb}
\paragraph{One-one adaptation.}
We reused the same baselines and assessed on a classification problem in the molecule domain: $\texttt{ogbg-molbace}$ and $\texttt{ogbg-molbbbp}$. In addition, multiple adaptation steps were enabled to observe the performance changes.
In particular, we trained GIN~\citep{xu2018gin} models on one dataset, then performed \gls{ttt} and evaluated on the test set of another. As the input and output features were categorical and the main task was \textit{graph-level binary classification}, we prepended an OGB-provided atom encoder to the shared encoder and appended a global addition readout to the main decoder. In addition, the main task was evaluated using ROC-AUC and relative gain w.r.t. the standard inference without any adaptation. 

For the \gls{ssl} decoder, we reused \textit{node masking} as the \gls{ssl} task, based on its stable loss signals and highest positive cosine similarity between task gradients (see Appendix~\ref{appendix:ssl_task_selection_ogb}). The auxiliary loss $\mathcal{L}_{ssl}$ was computed as the sum of cross-entropy losses between reconstructed and ``clean'' input atom features, whereas the main loss $\mathcal{L}_{main}$ was defined as a simple binary cross-entropy between predicted and true labels.

\begin{table*}[!h]
\centering
\small
\begin{adjustbox}{max width=\textwidth}
\begin{tabular}{ll|ccccc}
\hline
\multirow{2}{*}{\textbf{Adapt. Method}} & \multirow{2}{*}{\textbf{Step}} & \multicolumn{2}{c}{\texttt{ogbg-molbace$\rightarrow$ ogbg-molbbbp}} & \multicolumn{2}{c}{\texttt{ogbg-molbbbp $\rightarrow$ ogbg-molbace}} \\
 &  & \textbf{Test ROC-AUC$\uparrow$}& \textbf{Rel. Gain (\%)$\uparrow$} & \textbf{Test ROC-AUC$\uparrow$} & \textbf{Rel. Gain (\%)$\uparrow$} \\
\hline
\textbf{Tent} & $\text{0}^\dagger$  & 0.4621 $\pm$ 0.0586 & 0 & 0.4460 $\pm$ 0.2032 & 0 \\
     & 1  & 0.4682 $\pm$ 0.0646 & 1.32 & 0.4352 $\pm$ 0.2194 & -2.43 \\
     & 10 & 0.4678 $\pm$ 0.0641 & 1.23 & 0.4356 $\pm$ 0.2194 & -2.32 \\
     & 20 & 0.4677 $\pm$ 0.0640 & 1.21 & 0.4360 $\pm$ 0.2192 & -2.24 \\
     & 40 & 0.4678 $\pm$ 0.0640 & 1.23 & 0.4364 $\pm$ 0.2191 & -2.15 \\
\hline
\textbf{TTT} & $\text{0}^\ddagger$  & 0.4582 $\pm$ 0.0519 & 0 & 0.5228 $\pm$ 0.1538 & 0 \\
    & 1  & 0.4587 $\pm$ 0.0522 & 0.11 & 0.5294 $\pm$ 0.1420 & 1.26 \\
    & 10 & 0.4656 $\pm$ 0.0634 & 1.62 & 0.5399 $\pm$ 0.1308 & 3.28 \\
    & 20 & 0.4595 $\pm$ 0.0562 & 0.29 & 0.5366 $\pm$ 0.1326 & 2.64 \\
    & 40 & 0.4507 $\pm$ 0.0585 & -1.64 & 0.5192 $\pm$ 0.1550 & -0.69 \\
\hline
\textbf{TTT + Rotograd} & 0  & 0.4386 $\pm$ 0.0495 & 0 & 0.4352 $\pm$ 0.1420 & 0 \\
               & 1  & 0.4384 $\pm$ 0.0494 & -0.04 & 0.4421 $\pm$ 0.1480 & 1.60 \\
               & 10 & 0.4310 $\pm$ 0.0417 & -1.73 & 0.4882 $\pm$ 0.1464 & 12.19 \\
               & 20 & 0.4307 $\pm$ 0.0444 & -1.78 & 0.5260 $\pm$ 0.1384 & 20.86 \\
               & 40 & 0.4339 $\pm$ 0.0495 & -1.07 & 0.5411 $\pm$ 0.1302 & 24.34 \\
\hline
\textbf{T3R (ours)} & 0  & 0.4651 $\pm$ 0.0561 & 0 & 0.3941 $\pm$ 0.0755 & 0 \\
           & 1  & 0.5054 $\pm$ 0.0478 & 8.65 & 0.5234 $\pm$ 0.1274 & 32.81 \\
           & 10 & 0.5243 $\pm$ 0.0826 & 12.71 & \textbf{0.6682} $\pm$ 0.0217 & \textbf{69.56} \\
           & 20 & 0.5391 $\pm$ 0.0391 & 15.90 & 0.6248 $\pm$ 0.0797 & 58.53 \\
           & 40 & \textbf{0.5608} $\pm$ 0.0516 & \textbf{20.57} & 0.6420 $\pm$ 0.0667 & 62.90 \\
\hline
\multicolumn{5}{l}{$0^{\dagger}$ and $0^{\ddagger}$ indicate \gls{erm} and Joint Training without adaptation, respectively.}
\end{tabular}
\end{adjustbox}
\caption{\textbf{Cross-domain adaptation from} \texttt{ogbg-bace} \textbf{to} \texttt{ogbg-bbbp} \textbf{and vice versa.} Each method was evaluated over five independent runs. Careful tuning of the adaptation step led to greater relative improvement across methods.}
\label{tab:ogb_result}
\end{table*}

\paragraph{Results.} Table~\ref{tab:ogb_result} presents the performance of the \gls{t3r} approach evaluated after a fixed number of \gls{ttt} steps. Results are averaged over five independent runs. \gls{t3r} achieved the highest ROC-AUC scores, 0.5608, when adapting from $\texttt{ogbg-molbace}$ to $\texttt{ogbg-molbbbp}$, and 0.6682 in the reverse direction. In addition, the proposed approach yielded significant relative improvement of 8.65\%-69.56\% over standard inference without adaptation. These gains could be attributed to the weight updates of a partial main encoder. However, this improved accuracy comes at the cost of increased runtime due to the additional gradient updates, as shown in Appendix~\ref{appendix:relative_time}.

For other adaptation techniques, unlike in the regression setting, performance improved slightly on this classification task. Under longer adaptation, \text{Tent} tended to converge early, while \text{TTT + Rotograd} showed a roughly linear increase or decrease depending on the difficulty of the cross-domain shift. Similar to \gls{t3r}, \gls{ttt} required careful tuning of the adaptation step $t$ to maximize the benefit from adaptability. Otherwise, over-adaptation caused an abrupt drop in main-task performance. 

\subsection{Ablation studies}
\paragraph{General setup.}
The experiments followed an adaptation from \texttt{dwd-anytown} (1-year, smallest topology) to \texttt{dwd-exn} (24-hour, largest topology). We assessed the impact of hyperparameter shifts at test-time and layer-wise rotation matrices on \gls{t3r}. For additional visualizations, we refer the reader to Appendix~\ref{appendix:visualization}. 

\begin{figure}[!h]
    \centering
    \begin{subfigure}[b]{0.48\linewidth}
        \centering
        \includegraphics[width=\linewidth]{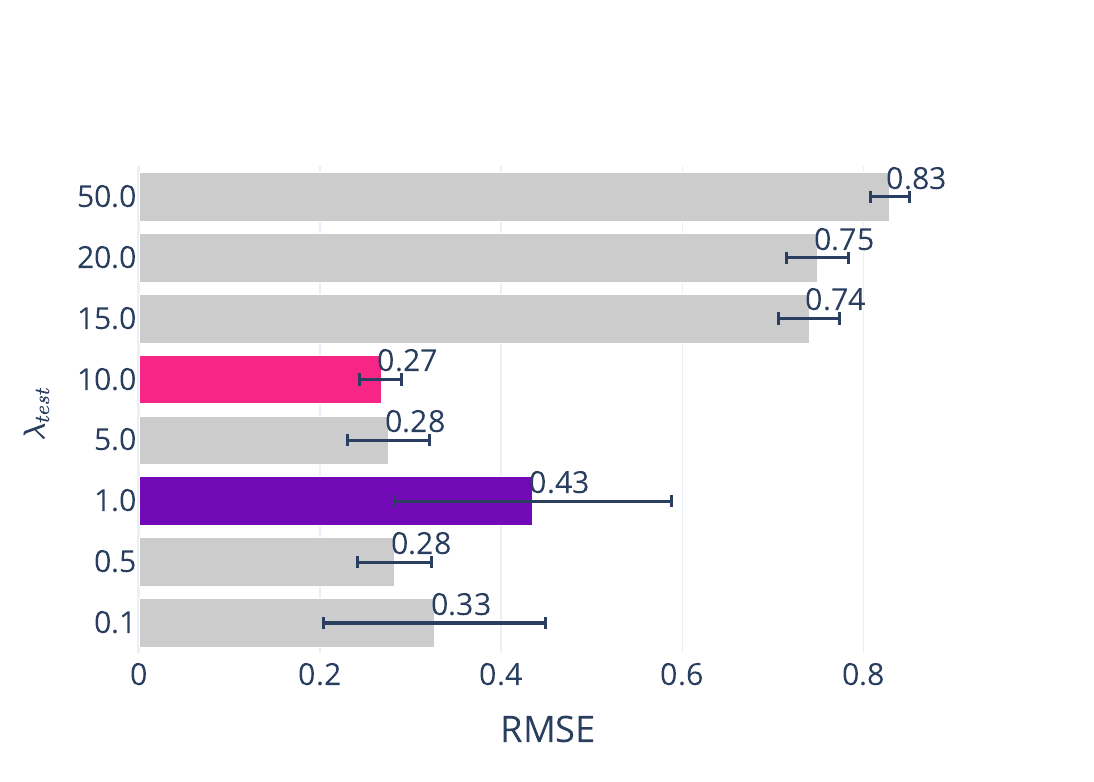}
        \caption{\textbf{Impact of the Auxiliary Loss Coefficient.} 
        }
        \label{fig:abla_lambda}
    \end{subfigure}
    \hfill
    \begin{subfigure}[b]{0.48\linewidth}
        \centering
        \includegraphics[width=\linewidth]{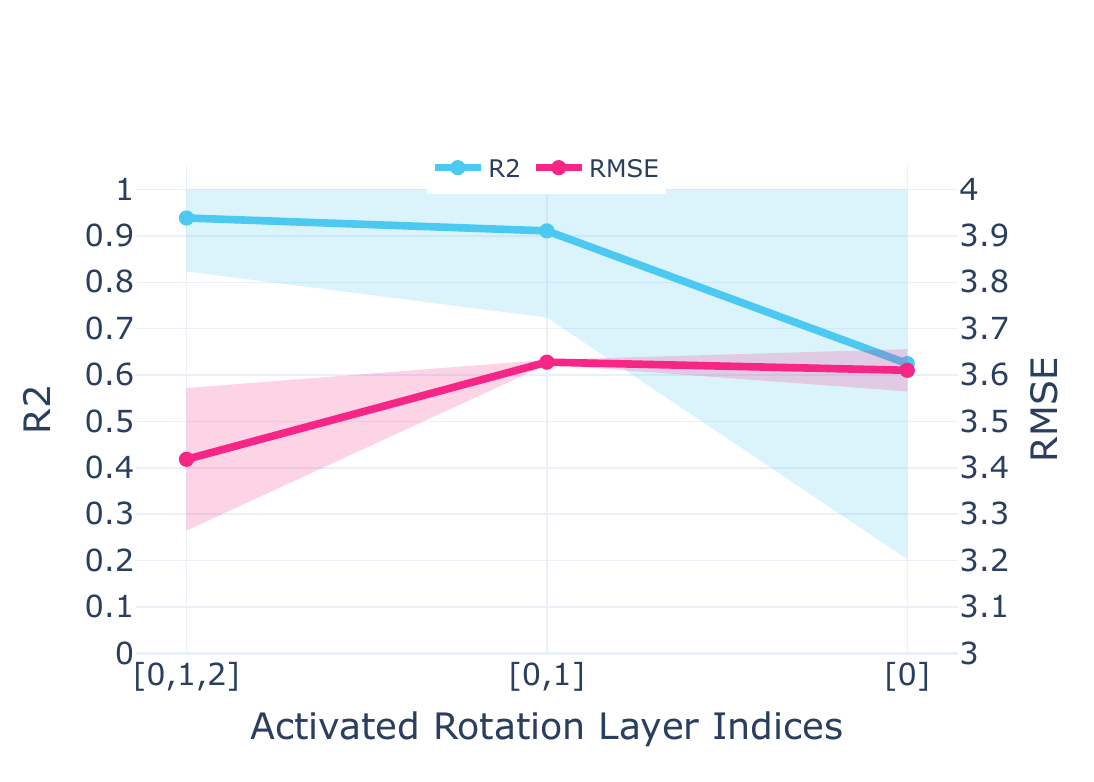}
        \caption{\textbf{Impact of rotation layers.}}
        \label{fig:abla_rotation_layers}
    \end{subfigure}
    
    \caption{\textbf{Ablation studies.} 
    Subfigure~\ref{fig:abla_lambda} illustrates the effect of the auxiliary-loss coefficient $\lambda_{test}$. The \textcolor[HTML]{7209B7}{purple} bar corresponds to the fixed setting without coefficient shifting (i.e., $\lambda_{train}=\lambda_{test}=1$), while the \textcolor[HTML]{F72585}{pink} bar highlights the best RMSE performance. Slight variations in $\lambda_{test}$ at inference tend to improve the target task. 
    Subfigure~\ref{fig:abla_rotation_layers} showcases the impact of rotation layers in T3R. Indices indicate the layers at which rotation matrices are attached, trained and test-time activated in the main and \gls{ssl} decoders. Removing rotation layers increases uncertainty and degrades main-task performance in terms of $\text{R}^2$ and RMSE.
    }
\end{figure}

\paragraph{The impact of auxiliary loss coefficient $\lambda$.} We fixed $\lambda_{\text{train}} = 1$ during training and varied $\lambda_{\text{test}}$ across multiple values at inference. Notably, this choice affected the \gls{ssl} signals from Equation~\ref{eqa:ttt_ssl} used to update the model’s weights.
Figure~\ref{fig:abla_lambda} shows the impact on RMSE, with similar trends observed across other metrics. Surprisingly, $\lambda_{\text{test}} = 10$ achieves the lowest RMSE. Compared to using the same value for \gls{ttt} and inference ($\lambda_{\text{test}} = \lambda_{\text{train}} = 1$), tuning $\lambda_{\text{test}}$ improves main task. However, tuning $\lambda$ to excessively large values leads to a decline in performance, likely due to over-adaptation.

\paragraph{The impact of layer-wise rotation layers.} We employed a default 3-layer GAT, sequentially ablated the Rotograd layers in both main and \gls{ssl} decoders, and reported their mean and std. values over 3 runs (Figure~\ref{fig:abla_rotation_layers}). Fewer rotation layers increased uncertainty and degraded $\text{R}^2$ and RMSE performance. 

\section{Conclusion}

We propose \gls{t3r}, a novel adaptation method on graphs that enhances affinity between the main and auxiliary tasks during training. At test-time training, \gls{t3r} applies a rotation-based transformation to redirect gradients from the auxiliary loss toward the main-task decoder, deeply adapting nearly the entire model without requiring access to ground-truth labels. Empirically, \gls{t3r} consistently outperforms standard baselines on both regression and classification datasets. These results highlight its potential for building robust, label-free adaptation pipelines in real-time graph-based systems, such as water networks, and in scenarios where fine-tuning or retraining is impractical.

\bibliography{bibliography}
\bibliographystyle{tmlr}

\appendix
\section{Appendix}

\subsection{Dataset Description}
\label{appendix:dataset_description}
Both \gls{dwd} and \gls{ogb} include several datasets served for a variety of tasks. However, only datasets with compatible inputs and identical tasks are suitable for model adaptation. Table~\ref{tab:dat_description} summarizes these datasets along with key statistics, as presented in Section~\ref{sec:collection}.

\begin{table}[H]
    \centering
    \begin{tabular}{l c c c}
        \hline
        \textbf{Dataset} & \textbf{$\#$records} & \textbf{$\#$nodes} & \textbf{$\#$edges} \\
        \hline
        \texttt{dwd-19pipe} & 673,846 & 12 & 21 \\
        \texttt{dwd-anytown} & 673,846 & 19 & 41 \\
        \texttt{dwd-jilin} & 673,846 & 27 & 34 \\
        \texttt{dwd-wa1} & 673,846 & 121  & 169 \\
        \texttt{dwd-npcl1} & 673,846 & 337 & 399 \\
        \texttt{dwd-ky7} & 673,846 & 481 & 604 \\
        \texttt{dwd-ky6} & 673,846 &  543 & 647 \\
        \texttt{dwd-ky4} & 673,846 &  959 & 1158 \\
        \texttt{dwd-ctown} & 1,846 &  388 & 444 \\
        \texttt{dwd-ky5} & 1,846 & 420 & 505 \\
        \texttt{dwd-ky13} & 1,846 & 778 & 944 \\
        \texttt{dwd-ky10} & 1,846 & 920  & 1061 \\
        \texttt{dwd-exn}  & 1,846 & 1891 & 2467 \\
        \hline
        \texttt{ogbg-molbace} & 2,039 & 24.1 & 51.9 \\
        \texttt{ogbg-molbbbp} & 1,513 & 34.1 & 73.7 \\
        \hline
    \end{tabular}
    \caption{\textbf{Summary of evaluation datasets.} For the water domain, we select eight networks simulated over one year and five networks simulated over 24 hours, where each record is processed as non-overlapping graph windows. In \gls{ogb}, each record is a unique molecule graph.}
    \label{tab:dat_description}
\end{table}

\subsection{\gls{ssl} Task Description}
\label{appendix:ssl_task_description}
We describe the graph-related \gls{ssl} tasks used in the Experiment~\ref{sec:ssl_task_selection}. 
\begin{itemize}
    \item \textbf{Attribute Masking}~\citep{feng2020grand,you2020GraphCL}: Reconstruct masked attributes from augmented graphs. We refer to this as \textit{node masking} (node-level) and \textit{edge masking} (edge-level).
    
    \item \textbf{\gls{mse} reconstruction in Embedding Space}: Instead of reconstructing raw attributes, we compute \gls{mse} loss between embeddings. This includes \textit{node emb MSE}, \textit{edge emb MSE}, and \textit{graph emb MSE}.
    
    \item \textbf{One-Sided Contrastive (Hinge) Loss}: To avoid the cost of full contrastive learning, we apply a Hinge loss that pulls augmented (positive) embeddings toward the clean (anchor) embedding. Named as \textit{\{node, edge, graph\} emb Hinge}, depending on the applied level.
    \item \textbf{One-Sided CLIP Loss}: Inspired by CLIP~\citep{radford2021LearningTV}, we treat graph indices as pseudo labels and train the SSL decoder to predict them using the inner-product of its output embedding vector. Denoted as \textit{\{node, edge, graph\} emb CLIP}.
    
    \item \textbf{Shortest Path Distance Regression}~\citep{li2020spd}: The SSL decoder regresses shortest path distances between nodes (computed via Dijkstra algorithm) based on edge weights. Referred to as \textit{edge SPD}.
\end{itemize}

\subsection{\gls{ssl} task selection on OGB}
\label{appendix:ssl_task_selection_ogb}

We re-ran the same task selection experiment but on the OGB benchmark (Section~\ref{sec:gpp_on_ogb}). In particular, we trained a GIN model on \texttt{ogbg-molbbbp} and performed \gls{ttt} on \texttt{ogbg-molbace}. Table~\ref{tab:ssl_selection_ogb} confirms a consistent observation: \textit{node masking} remains the top-performing \gls{ssl} objective with the strongest train gradient alignment.

\begin{table}[H]
    \centering
    \small
    \begin{tabular}{l|cccc}
    \hline
    \textbf{\gls{ssl} Task} & \textbf{\makecell{Test\\MAE(t=0)$\downarrow$} } & \textbf{\makecell{Test\\ MAE(t=1)$\downarrow$}} & \textbf{\makecell{Relative \\ Gain(\%)$\uparrow$}}&
    \textbf{\makecell{Train\\Gradient Alignment}} 
    \\
    \hline
    \textbf{Node mask }       & 0.51 & 0.56 & \textbf{10.07}  & \textbf{0.10} \\
    \textbf{Node emb CLIP}    & \textbf{0.31 }&\textbf{ 0.26} & -15.94 & 0.07 \\
    \textbf{Node emb Hinge}   & 0.46 & 0.47 & 1.43   & 0.02 \\
    \textbf{Graph emb CLIP}   & 0.41 & 0.41 & 0.35   & 0.02 \\
    \textbf{Graph emb MSE}    & 0.52 & 0.51 & -2.83  & 0.00 \\
    \textbf{Node emb MSE}     & 0.48 & 0.49 & 1.32   & 0.00 \\
    \textbf{Graph emb Hinge}  & 0.42 & 0.34 & -19.17 & -0.03 \\
    \hline
    \end{tabular}
    \caption{\textbf{\gls{ssl} Task Selection on OGB with GIN.} Note that GIN does not support edge attributes, so edge-related \gls{ssl} tasks are excluded.}  
    \label{tab:ssl_selection_ogb} 
\end{table}

\subsection{Hyperparameter setting}
\label{appendix:hyper_tuning}
Table~\ref{tab:hyper_parameters} reports the optimal hyperparameters for \gls{t3r} and other baselines introduced in Section~\ref{sec:nsp_on_dwd}. Note that baselines use different normalization types: \gls{erm} and \text{Tent} employ BatchNorm, whereas \gls{ttt} and \text{\gls{ttt} + Rotograd} use LayerNorm for both tasks, as this empirically yields better performance. 

\begin{table}[H]
    \centering
    \small
    \begin{tabular}{lcc}
        \hline
        \textbf{Hyperparameter} & \textbf{NSP} & \textbf{GPP} \\
        \hline
        Graph Conv. & GAT &GIN\\ 
        $\#$layers & 3 & 3\\ 
        $\#$ hidden channels & 64 &  300\\
        Norm type & z-score & - \\
        Normalization layer & LayerNorm & BatchNorm\\
        Main Readout & - & Global add pool \\
        Main loss & MSE & BCE\\
        \gls{ssl} loss & Node mask & Sum of CE losses\\
        Epochs & 30 & 50 \\
        Batch size & 128 & 480 \\
        Train $\mid$ TTT weight decay & 1e-5 $\mid$ 0. & 1e-6 $\mid$ 1e-6\\
        Train $\mid$ TTT learning rate &  5e-3 $\mid$ 5e-4 & 0.01 $\mid$ 1e-3 \\
        Train $\mid$ TTT Aug. NodeDrop &  0.95 $\mid$ 0.1 & 0.95 $\mid$  0.1\\
        Train $\mid$ TTT Aug. EdgeDrop &  0.1$\mid$   0.1 & 0.1 $\mid$  0.1\\
        Train $\mid$ TTT coefficient $\lambda$ & 1.0 $\mid$  10. & 1.0 $\mid$  1.0\\
        Dropout rate & 0.5 & 0.5\\
        Activation & ReLU & ReLU \\
        Optimizer & Adam & Adam \\
        Scheduler & Cosine Annealing & - \\
        \hline

    \end{tabular}
    \caption{\textbf{Hyperparameter settings for the Next-State Prediction (NSP) and Graph Property Prediction (GPP) tasks.}}
    \label{tab:hyper_parameters}
\end{table}

\subsection{Relative time measurement}
\label{appendix:relative_time}
Extending layer-wise rotation matrices in GNNs increases computational cost. We report the computation overhead of adaptation methods on OGB in Section~\ref{sec:gpp_on_ogb}. In particular, T3R is 1.35x-9.35x slower than ERM at inference, depending on the adaptation step (t) (Table~\ref{tab:rel_time}). The trade-off seems acceptable given the substantial performance gains (+8.65\% to +69.56\% on OGB) and the fact that GNNs often perform best with shallow architectures. In addition, this accuracy-efficiency trade-off can be managed by tuning t; better improvements are often achieved in the first step (t=1).

\begin{table}[H]
\centering
\begin{tabular}{lccccc}
\hline
&\multicolumn{5}{c}{\textbf{Relative runtime in seconds (slowdown relative to ERM)}}\\
\hline
\textbf{Method} & \textbf{t=0} &\textbf{ t=1 }& \textbf{t=10} & \textbf{t=20} & \textbf{t=40} \\
\hline
\textbf{Tent} & \textbf{2.99 (1.00x)} & \textbf{3.47 (1.16x)} & \textbf{4.80 (1.61x)} & \textbf{6.04 (2.02x)} & \textbf{8.72 (2.91x)} \\
\textbf{TTT} & 3.29 (1.10x) & 3.75 (1.25x) & 7.20 (2.40x) & 9.99 (3.34x) & 15.54 (5.19x) \\
\textbf{TTT+Rotograd} & 3.50 (1.17x) & 4.10 (1.37x) & 7.90 (2.64x) & 11.32 (3.78x) & 17.80 (5.95x) \\
\textbf{T3R (Ours)} & 4.05 (1.35x) & 5.24 (1.75x) & 12.74 (4.25x) & 19.45 (6.50x) & 28.27 (9.45x) \\
\hline
\end{tabular}
\caption{\textbf{Computation overhead on OGB}. $t$ denotes the number of adaptation steps. Results may vary depending on the hardware configuration. We conducted this experiment on a local NVIDIA GeForce RTX 3060 Laptop GPU (6GB VRAM).}
\label{tab:rel_time}
\end{table}

\subsection{Per-dataset Next-State Prediction results}
\label{appendix:per_dataset_results}

The full results of Experiment~\ref{exp:per_benchmark_nsp} across five metrics for each water-related dataset are presented in the following tables:
\begin{table}[H]
\centering
\small
\renewcommand{\arraystretch}{1.2}
\begin{tabular}{l|ccccc}
\hline
\textbf{Method} & \textbf{R2$\uparrow$} & \textbf{RMSE$\downarrow$} & \textbf{PCC$\uparrow$} & \textbf{MAE$\downarrow$} & \textbf{NSE$\uparrow$} \\
\hline
\textbf{ERM} & 0.9297 $\pm$ 0.0030 & 0.5844 $\pm$ 0.0303 & 0.9623 $\pm$ 0.0019 & 0.4284 $\pm$ 0.0198 & -11.8799 $\pm$ 1.2133 \\
\textbf{Joint Training} & 0.9405 $\pm$ 0.0018 & \textbf{0.4137} $\pm$ 0.0077 & 0.9687 $\pm$ 0.0011 & \textbf{0.2655} $\pm$ 0.0117 & -10.6401 $\pm$ 2.3255 \\
\hline
\textbf{Tent} & 0.9279 $\pm$ 0.0013 & 0.6027 $\pm$ 0.0143 & 0.9611 $\pm$ 0.0008 & 0.4196 $\pm$ 0.0108 & -12.9159 $\pm$ 0.6678 \\
\textbf{TTT} & 0.9408 $\pm$ 0.0017 & 0.4169 $\pm$ 0.0125 & 0.9689 $\pm$ 0.0011 & 0.2681 $\pm$ 0.0135 & -10.4934 $\pm$ 2.0832 \\
\textbf{TTT + Rotograd} & \textbf{0.9427} $\pm$ 0.0109 & 0.4420 $\pm$ 0.2380 & \textbf{0.9699} $\pm$ 0.0062 & 0.3600 $\pm$ 0.2704 & \textbf{-5.0650} $\pm$ 4.8243 \\
\textbf{T3R (ours)} & 0.9352 $\pm$ 0.0104 & 0.5966 $\pm$ 0.0177 & 0.9656 $\pm$ 0.0062 & 0.4348 $\pm$ 0.0168 & -13.8653 $\pm$ 1.9707 \\
\hline
\end{tabular} 
\caption{\textbf{Results for next-state prediction on} \texttt{dwd-wa1}. Values show mean $\pm$ standard deviation over multiple runs. Best-performing metrics in each column are highlighted in bold.}
\end{table}

\begin{table}[h]
\centering
\small
\renewcommand{\arraystretch}{1.2}
\begin{tabular}{l|ccccc}
\hline
\textbf{Method} & \textbf{R2$\uparrow$} & \textbf{RMSE$\downarrow$} & \textbf{PCC$\uparrow$} & \textbf{MAE$\downarrow$} & \textbf{NSE$\uparrow$} \\
\hline
\textbf{ERM} & 0.7821 $\pm$ 0.0419 & 0.5097 $\pm$ 0.0369 & 0.8839 $\pm$ 0.0238 & 0.4224 $\pm$ 0.0424 & 0.7070 $\pm$ 0.0410 \\
\textbf{Joint Training} & 0.8510 $\pm$ 0.0232 & 0.4708 $\pm$ 0.0885 & 0.9223 $\pm$ 0.0126 & 0.3566 $\pm$ 0.1318 & 0.7416 $\pm$ 0.0905 \\
\hline
\textbf{Tent} & 0.7896 $\pm$ 0.0260 & 0.4989 $\pm$ 0.0553 & 0.8883 $\pm$ 0.0146 & 0.4144 $\pm$ 0.0658 & 0.7175 $\pm$ 0.0596 \\
\textbf{TTT} & 0.8509 $\pm$ 0.0234 & 0.4707 $\pm$ 0.0881 & 0.9223 $\pm$ 0.0127 & 0.3565 $\pm$ 0.1322 & 0.7417 $\pm$ 0.0902 \\
\textbf{TTT + Rotograd} & 0.8222 $\pm$ 0.0195 & 0.6235 $\pm$ 0.1246 & 0.9066 $\pm$ 0.0108 & 0.4983 $\pm$ 0.0093 & 0.5447 $\pm$ 0.1925 \\
\textbf{T3R (ours)} & \textbf{0.9724} $\pm$ 0.0003 & \textbf{0.1661} $\pm$ 0.0047 & \textbf{0.9861} $\pm$ 0.0001 & \textbf{0.1042} $\pm$ 0.0068 & \textbf{0.9684} $\pm$ 0.0022 \\
\hline
\end{tabular} 
\caption{\textbf{Results for next-state prediction on} \texttt{dwd-ky5}. Values show mean $\pm$ standard deviation over multiple runs. Best-performing metrics in each column are highlighted in bold.}
\end{table}

\begin{table}[h]
\centering
\small
\renewcommand{\arraystretch}{1.2}
\begin{tabular}{l|ccccc}
\hline
\textbf{Method} & \textbf{R2$\uparrow$} & \textbf{RMSE$\downarrow$} & \textbf{PCC$\uparrow$} & \textbf{MAE$\downarrow$} & \textbf{NSE$\uparrow$} \\
\hline
\textbf{ERM} & 0.9698 $\pm$ 0.0038 & 0.6459 $\pm$ 0.0104 & 0.9842 $\pm$ 0.0022 & 0.4487 $\pm$ 0.0068 & 0.3321 $\pm$ 0.2544 \\
\textbf{Joint Training} & \textbf{0.9798} $\pm$ 0.0058 & 0.4014 $\pm$ 0.1729 & 0.9896 $\pm$ 0.0030 & 0.2888 $\pm$ 0.1360 & 0.5926 $\pm$ 0.1483 \\
\hline
\textbf{Tent} & 0.9706 $\pm$ 0.0032 & 0.6282 $\pm$ 0.0245 & 0.9846 $\pm$ 0.0018 & 0.4354 $\pm$ 0.0176 & 0.3302 $\pm$ 0.2299 \\
\textbf{TTT} & \textbf{0.9798} $\pm$ 0.0058 & 0.4116 $\pm$ 0.1869 & \textbf{0.9897} $\pm$ 0.0030 & 0.2967 $\pm$ 0.1452 & 0.5923 $\pm$ 0.1707 \\
\textbf{TTT + Rotograd} & 0.9716 $\pm$ 0.0157 & \textbf{0.3417} $\pm$ 0.1092 & 0.9852 $\pm$ 0.0085 & \textbf{0.2679} $\pm$ 0.1038 & \textbf{0.8086} $\pm$ 0.1464 \\
\textbf{T3R (ours)} & 0.9700 $\pm$ 0.0090 & 0.3783 $\pm$ 0.0705 & 0.9843 $\pm$ 0.0050 & 0.2867 $\pm$ 0.0570 & 0.7010 $\pm$ 0.1479 \\
\hline
\end{tabular} 
\caption{\textbf{Results for next-state prediction on} \texttt{dwd-ky6}. A single adaptation step is applied for all methods except the first two (no-adaptation baselines). Values show mean $\pm$ standard deviation over three runs. Best-performing metrics are highlighted in bold.}
\end{table}

\begin{table}[h]
\centering
\small
\renewcommand{\arraystretch}{1.2}
\begin{tabular}{l|ccccc}
\hline
\textbf{Method} & \textbf{R2$\uparrow$} & \textbf{RMSE$\downarrow$} & \textbf{PCC$\uparrow$} & \textbf{MAE$\downarrow$} & \textbf{NSE$\uparrow$} \\
\hline
\textbf{ERM} & 0.8300 $\pm$ 0.0211 & 0.5219 $\pm$ 0.0324 & 0.9109 $\pm$ 0.0116 & 0.4689 $\pm$ 0.0363 & 0.6922 $\pm$ 0.0383 \\
\textbf{Joint Training} & 0.7989 $\pm$ 0.1035 & 0.5441 $\pm$ 0.2618 & 0.8925 $\pm$ 0.0572 & 0.4340 $\pm$ 0.2570 & 0.6127 $\pm$ 0.3231 \\
\hline
\textbf{Tent} & 0.7677 $\pm$ 0.0512 & 0.5292 $\pm$ 0.0387 & 0.8757 $\pm$ 0.0295 & 0.4406 $\pm$ 0.0444 & 0.6832 $\pm$ 0.0459 \\
\textbf{TTT} & 0.7973 $\pm$ 0.1053 & 0.5449 $\pm$ 0.2622 & 0.8916 $\pm$ 0.0583 & 0.4334 $\pm$ 0.2570 & 0.6116 $\pm$ 0.3241 \\
\textbf{TTT + Rotograd} & 0.7406 $\pm$ 0.0602 & 0.8212 $\pm$ 0.1643 & 0.8600 $\pm$ 0.0354 & 0.6914 $\pm$ 0.2368 & 0.2184 $\pm$ 0.2873 \\
\textbf{T3R (ours)} & \textbf{0.9760} $\pm$ 0.0089 & \textbf{0.1699} $\pm$ 0.0469 & \textbf{0.9879} $\pm$ 0.0045 & \textbf{0.1270} $\pm$ 0.0514 & \textbf{0.9656} $\pm$ 0.0195 \\
\hline
\end{tabular} 
\caption{\textbf{Results for next-state prediction on} \texttt{dwd-ky13}. Values show mean $\pm$ standard deviation over multiple runs. Best-performing metrics in each column are highlighted in bold.}
\end{table}

\begin{table}[H]
\centering
\small
\renewcommand{\arraystretch}{1.2}
\begin{tabular}{l|ccccc}
\hline
\textbf{Method} & \textbf{R2$\uparrow$} & \textbf{RMSE$\downarrow$} & \textbf{PCC$\uparrow$} & \textbf{MAE$\downarrow$} & \textbf{NSE$\uparrow$} \\
\hline
\textbf{ERM} & 0.9429 $\pm$ 0.0036 & 0.4529 $\pm$ 0.0511 & 0.9695 $\pm$ 0.0021 & 0.3679 $\pm$ 0.0394 & -0.2282 $\pm$ 0.0399 \\
\textbf{Joint Training} & 0.9608 $\pm$ 0.0017 & 0.3436 $\pm$ 0.0504 & 0.9798 $\pm$ 0.0009 & 0.2430 $\pm$ 0.0356 & -0.0033 $\pm$ 0.3232 \\
\hline
\textbf{Tent} & 0.9407 $\pm$ 0.0030 & 0.4723 $\pm$ 0.0243 & 0.9682 $\pm$ 0.0018 & 0.3715 $\pm$ 0.0287 & -0.3173 $\pm$ 0.1090 \\
\textbf{TTT} & \textbf{0.9608} $\pm$ 0.0018 & 0.3552 $\pm$ 0.0508 & \textbf{0.9798} $\pm$ 0.0009 & 0.2521 $\pm$ 0.0361 & 0.0027 $\pm$ 0.3227 \\
\textbf{TTT + Rotograd} & 0.9234 $\pm$ 0.0629 & \textbf{0.2326} $\pm$ 0.0273 & 0.9053 $\pm$ 0.1273 & \textbf{0.1539} $\pm$ 0.0309 & \textbf{0.6495} $\pm$ 0.0267 \\
\textbf{T3R (ours)} & 0.9461 $\pm$ 0.0312 & 0.3772 $\pm$ 0.1163 & 0.9722 $\pm$ 0.0161 & 0.2745 $\pm$ 0.0802 & -0.3626 $\pm$ 1.7306 \\
\hline
\end{tabular} 
\caption{\textbf{Results for next-state prediction on} \texttt{dwd-npcl1}. Values show mean $\pm$ standard deviation over multiple runs. Best-performing metrics in each column are highlighted in bold.}
\end{table}

\begin{table}[H]
\centering
\small
\renewcommand{\arraystretch}{1.2}
\begin{tabular}{l|ccccc}
\hline
\textbf{Method} & \textbf{R2$\uparrow$} & \textbf{RMSE$\downarrow$} & \textbf{PCC$\uparrow$} & \textbf{MAE$\downarrow$} & \textbf{NSE$\uparrow$} \\
\hline
\textbf{ERM} & 0.9717 $\pm$ 0.0102 & 0.3901 $\pm$ 0.1600 & 0.9852 $\pm$ 0.0056 & 0.3268 $\pm$ 0.1645 & 0.1836 $\pm$ 0.6739 \\
\textbf{Joint Training} & \textbf{0.9776} $\pm$ 0.0071 & \textbf{0.3128} $\pm$ 0.0910 & 0.9885 $\pm$ 0.0038 & \textbf{0.2302} $\pm$ 0.0878 & -0.1863 $\pm$ 0.7491 \\
\hline
\textbf{Tent} & 0.9693 $\pm$ 0.0106 & 0.3912 $\pm$ 0.1224 & 0.9839 $\pm$ 0.0058 & 0.3185 $\pm$ 0.1315 & -0.0647 $\pm$ 0.5163 \\
\textbf{TTT} & 0.9778 $\pm$ 0.0073 & 0.3148 $\pm$ 0.0988 & \textbf{0.9886} $\pm$ 0.0039 & 0.2318 $\pm$ 0.0929 & -0.1858 $\pm$ 0.7592 \\
\textbf{TTT + Rotograd} & 0.9748 $\pm$ 0.0042 & 0.4092 $\pm$ 0.1704 & 0.9872 $\pm$ 0.0022 & 0.3264 $\pm$ 0.1423 & \textbf{0.5820} $\pm$ 0.3215 \\
\textbf{T3R (ours)} & 0.8747 $\pm$ 0.1670 & 0.3345 $\pm$ 0.1259 & 0.9194 $\pm$ 0.1140 & 0.2538 $\pm$ 0.0948 & 0.5359 $\pm$ 0.4260 \\
\hline
\end{tabular} 
\caption{\textbf{Results for next-state prediction on} \texttt{dwd-ky7}. Values show mean $\pm$ standard deviation over multiple runs. Best-performing metrics in each column are highlighted in bold.}
\end{table}

\begin{table}[H]
\centering
\small
\renewcommand{\arraystretch}{1.2}
\begin{tabular}{l|ccccc}
\hline
\textbf{Method} & \textbf{R2$\uparrow$} & \textbf{RMSE$\downarrow$} & \textbf{PCC$\uparrow$} & \textbf{MAE$\downarrow$} & \textbf{NSE$\uparrow$} \\
\hline
\textbf{ERM} & \textbf{0.9778} $\pm$ 0.0120 & 0.3130 $\pm$ 0.0722 & \textbf{0.9885} $\pm$ 0.0064 & \textbf{0.2307} $\pm$ 0.0593 & 0.1304 $\pm$ 0.8040 \\
\textbf{Joint Training} & 0.9720 $\pm$ 0.0054 & 0.4917 $\pm$ 0.1514 & 0.9857 $\pm$ 0.0028 & 0.4076 $\pm$ 0.1192 & -1.4581 $\pm$ 1.4893 \\
\hline
\textbf{Tent} & 0.9737 $\pm$ 0.0116 & 0.3254 $\pm$ 0.0897 & 0.9863 $\pm$ 0.0063 & 0.2382 $\pm$ 0.0682 & -0.0816 $\pm$ 0.8438 \\
\textbf{TTT} & 0.9724 $\pm$ 0.0057 & 0.4888 $\pm$ 0.1465 & 0.9859 $\pm$ 0.0029 & 0.4061 $\pm$ 0.1176 & -1.3802 $\pm$ 1.4800 \\
\textbf{TTT + Rotograd} & 0.9098 $\pm$ 0.0851 & \textbf{0.2791 }$\pm$ 0.0402 & 0.9366 $\pm$ 0.0718 & 0.2422 $\pm$ 0.0473 & \textbf{0.3661 }$\pm$ 0.4753 \\
\textbf{T3R (ours)} & 0.9755 $\pm$ 0.0063 & 0.3070 $\pm$ 0.2043 & 0.9874 $\pm$ 0.0034 & 0.2309 $\pm$ 0.1476 & 0.2824 $\pm$ 0.9126 \\
\hline
\end{tabular} 
\caption{\textbf{Results for next-state prediction on} \texttt{dwd-ky4}. Values show mean $\pm$ standard deviation over multiple runs. Best-performing metrics in each column are highlighted in bold.}
\end{table}

\begin{table}[H]
\centering
\small
\renewcommand{\arraystretch}{1.2}
\begin{tabular}{l|ccccc}
\hline
\textbf{Method} & \textbf{R2$\uparrow$} & \textbf{RMSE$\downarrow$} & \textbf{PCC$\uparrow$} & \textbf{MAE$\downarrow$} & \textbf{NSE$\uparrow$} \\
\hline
\textbf{ERM} & 0.8122 $\pm$ 0.0819 & 0.5186 $\pm$ 0.0150 & 0.9004 $\pm$ 0.0455 & 0.4024 $\pm$ 0.0694 & 0.7024 $\pm$ 0.0169 \\
\textbf{Joint Training} & 0.7834 $\pm$ 0.0342 & 0.6189 $\pm$ 0.2543 & 0.8849 $\pm$ 0.0194 & 0.4911 $\pm$ 0.2911 & 0.5281 $\pm$ 0.3830 \\
\hline
\textbf{Tent} & 0.8187 $\pm$ 0.0746 & 0.5705 $\pm$ 0.0769 & 0.9041 $\pm$ 0.0412 & 0.4653 $\pm$ 0.1311 & 0.6359 $\pm$ 0.0934 \\
\textbf{TTT} & 0.7850 $\pm$ 0.0360 & 0.6173 $\pm$ 0.2568 & 0.8857 $\pm$ 0.0204 & 0.4899 $\pm$ 0.2928 & 0.5294 $\pm$ 0.3853 \\
\textbf{TTT + Rotograd} & 0.7008 $\pm$ 0.0911 & 0.7632 $\pm$ 0.1331 & 0.8357 $\pm$ 0.0556 & 0.5849 $\pm$ 0.0089 & 0.3435 $\pm$ 0.2332 \\
\textbf{T3R (ours)} & \textbf{0.9672} $\pm$ 0.0059 & \textbf{0.1782} $\pm$ 0.0140 & \textbf{0.9835} $\pm$ 0.0030 & \textbf{0.1183} $\pm$ 0.0093 & \textbf{0.9647} $\pm$ 0.0054 \\
\hline
\end{tabular} 
\caption{\textbf{Results for next-state prediction on} \texttt{dwd-ky10}. Values show mean $\pm$ standard deviation over multiple runs. Best-performing metrics in each column are highlighted in bold.}
\end{table}

\begin{table}[H]
\centering
\small
\renewcommand{\arraystretch}{1.2}
\begin{tabular}{l|ccccc}
\hline
\textbf{Method} & \textbf{R2$\uparrow$} & \textbf{RMSE$\downarrow$} & \textbf{PCC$\uparrow$} & \textbf{MAE$\downarrow$} & \textbf{NSE$\uparrow$} \\
\hline
\textbf{ERM} & \textbf{0.9297} $\pm$ 0.0030 & 0.5844 $\pm$ 0.0303 & \textbf{0.9623} $\pm$ 0.0019 & 0.4284 $\pm$ 0.0198 & -11.8800 $\pm$ 1.2133 \\
\textbf{Joint Training} & 0.6836 $\pm$ 0.0568 & 0.7063 $\pm$ 0.2614 & 0.8263 $\pm$ 0.0348 & 0.5417 $\pm$ 0.2413 & 0.4538 $\pm$ 0.4100 \\
\hline
\textbf{Tent} & 0.9279 $\pm$ 0.0013 & 0.6027 $\pm$ 0.0143 & 0.9611 $\pm$ 0.0008 & 0.4196 $\pm$ 0.0108 & -12.9159 $\pm$ 0.6678 \\
\textbf{TTT} & 0.6837 $\pm$ 0.0568 & 0.7054 $\pm$ 0.2599 & 0.8264 $\pm$ 0.0347 & 0.5411 $\pm$ 0.2400 & 0.4556 $\pm$ 0.4070 \\
\textbf{TTT + Rotograd} & 0.7008 $\pm$ 0.0911 & 0.7632 $\pm$ 0.1331 & 0.8357 $\pm$ 0.0556 & 0.5849 $\pm$ 0.0089 & 0.3435 $\pm$ 0.2332 \\
\textbf{T3R (ours)} & 0.9258 $\pm$ 0.0099 & \textbf{0.2778} $\pm$ 0.0136 & 0.9622 $\pm$ 0.0052 & \textbf{0.1463} $\pm$ 0.0224 & \textbf{0.9225} $\pm$ 0.0076 \\
\hline
\end{tabular} 
\caption{\textbf{Results for next-state prediction on} \texttt{dwd-ctown}. Values show mean $\pm$ standard deviation over multiple runs. Best-performing metrics in each column are highlighted in bold.}
\end{table}

\subsection{Visualization}
\label{appendix:visualization}
\begin{figure}[H]
    \centering
    \begin{subfigure}{0.48\textwidth}
        \centering
        \includegraphics[width=\linewidth]{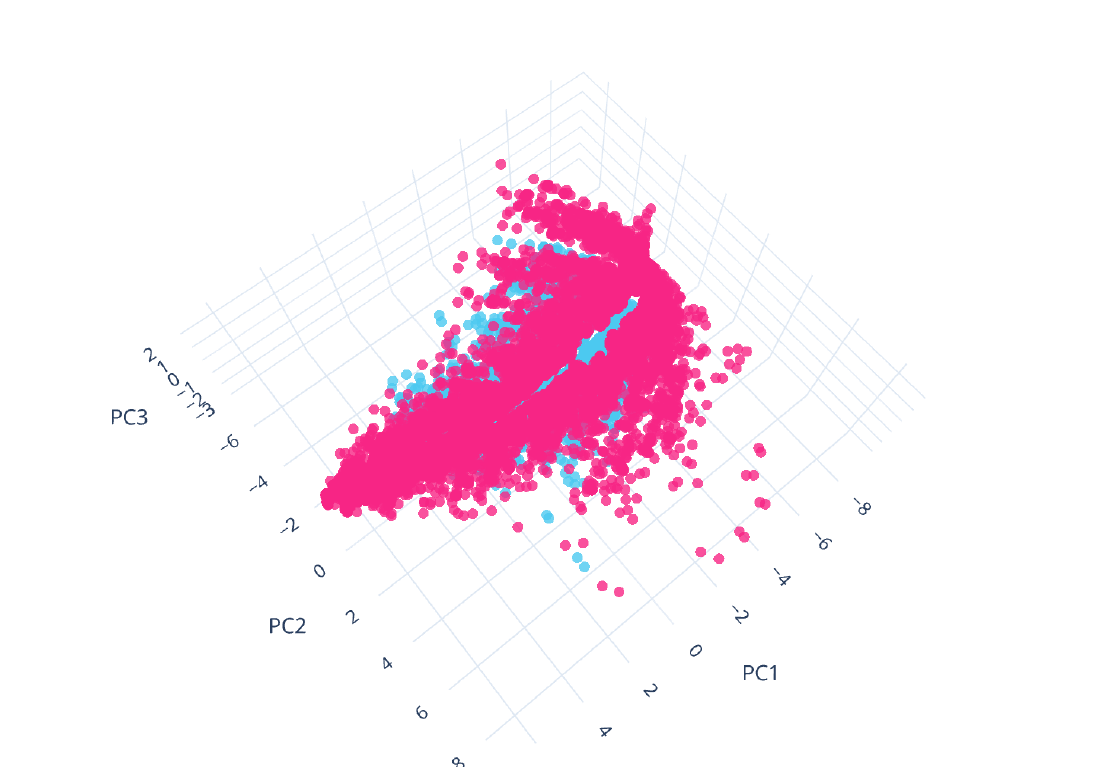}
        \caption{TTT- Encoder's last layer emb.}
    \end{subfigure}
    \hfill
    \begin{subfigure}{0.48\textwidth}
        \centering
        \includegraphics[width=\linewidth]{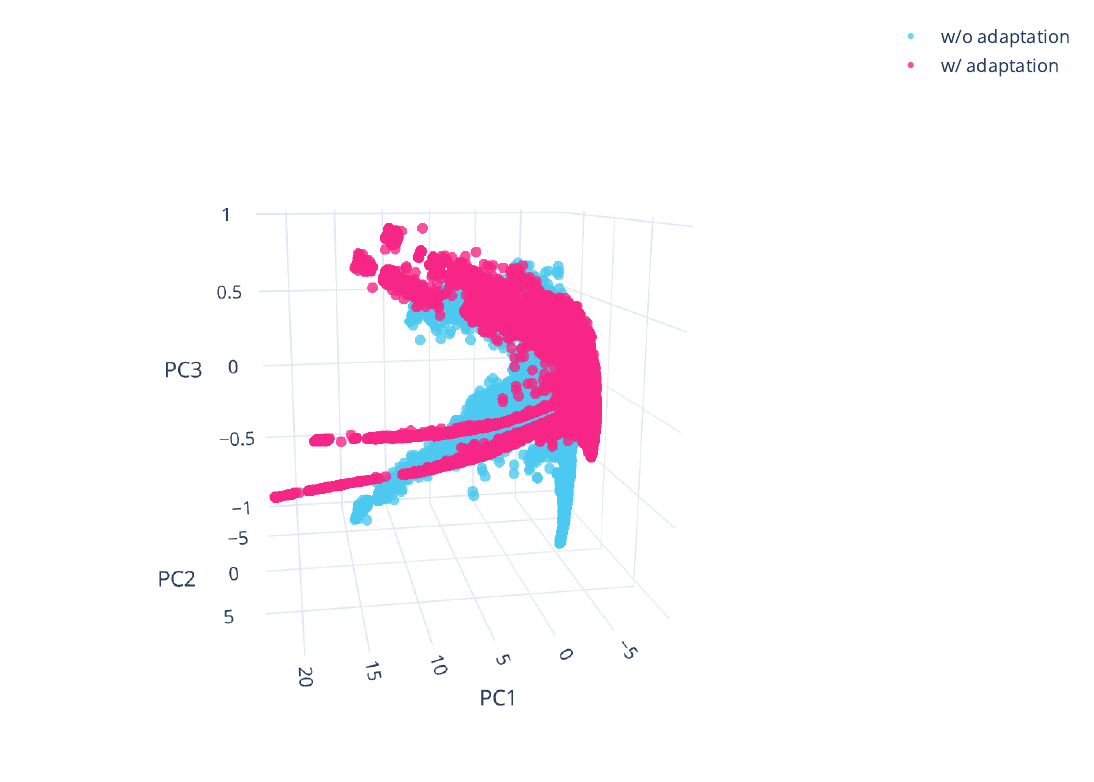}
        \caption{T3R- Encoder's last layer emb.}
    \end{subfigure}


    \begin{subfigure}{0.48\textwidth}
        \centering
        \includegraphics[width=\linewidth]{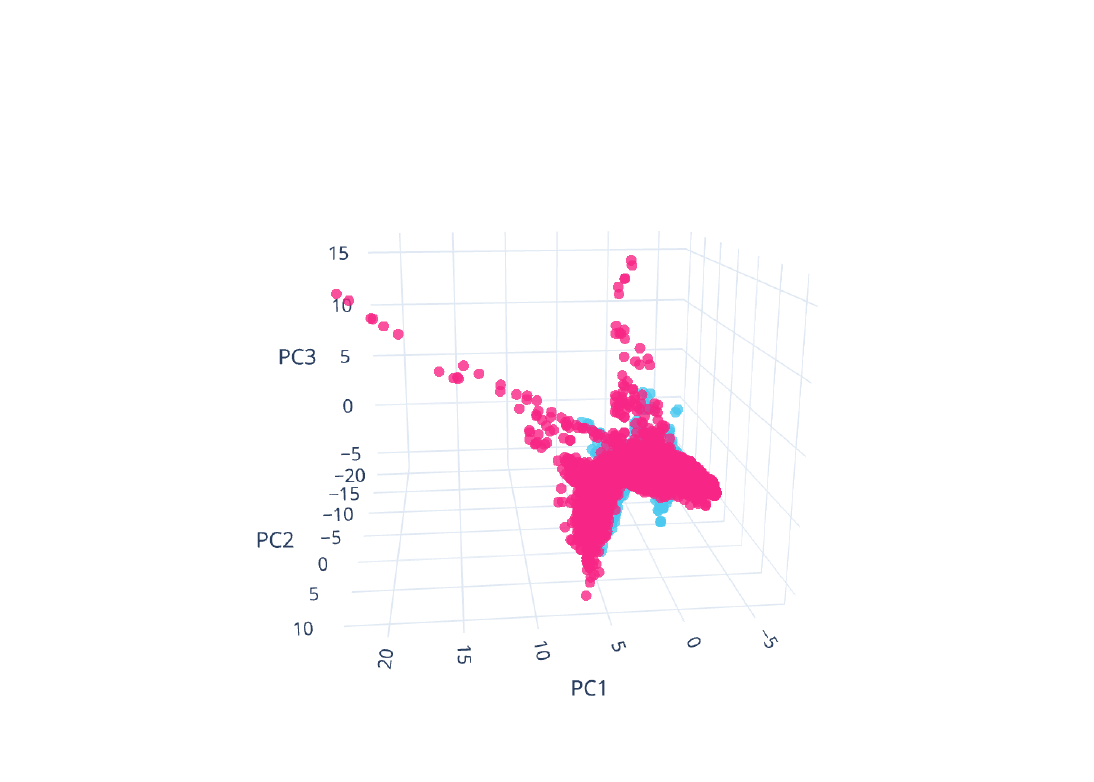}
        \caption{TTT- Main decoder's penultimate emb.}
    \end{subfigure}
    \hfill
    \begin{subfigure}{0.48\textwidth}
        \centering
        \includegraphics[width=\linewidth]{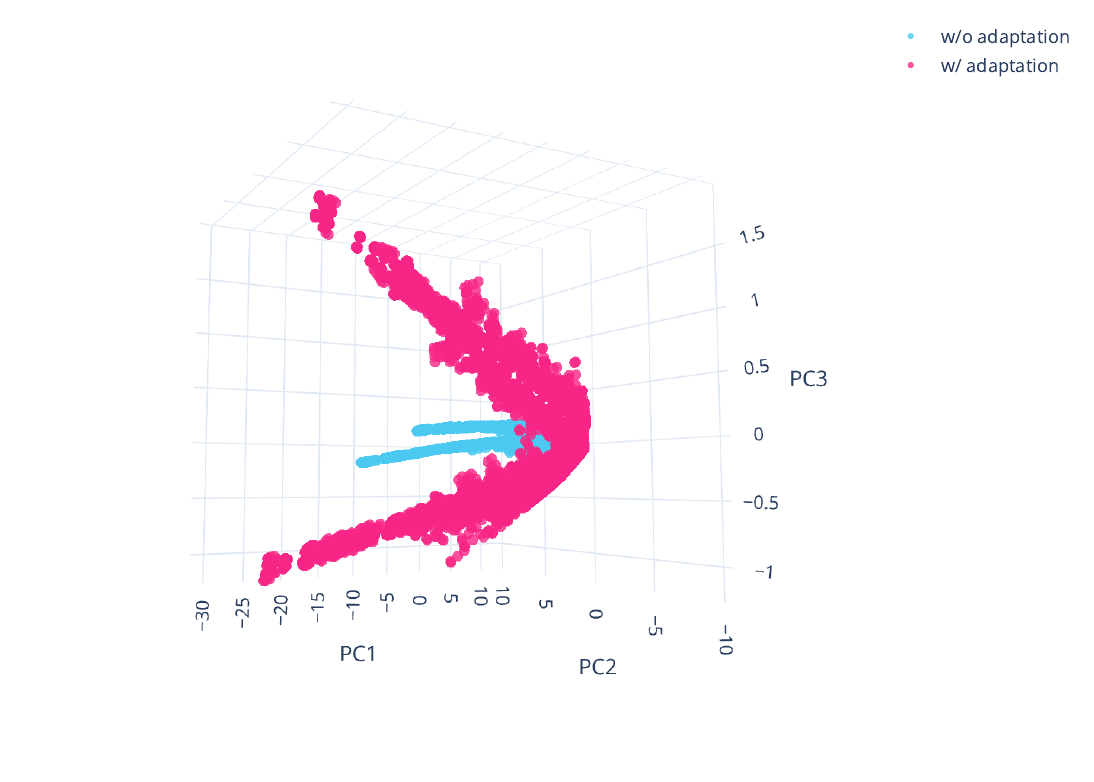}
        \caption{T3R- Main decoder's penultimate emb.}
    \end{subfigure}


    \begin{subfigure}{0.48\textwidth}
        \centering
        \includegraphics[width=\linewidth]{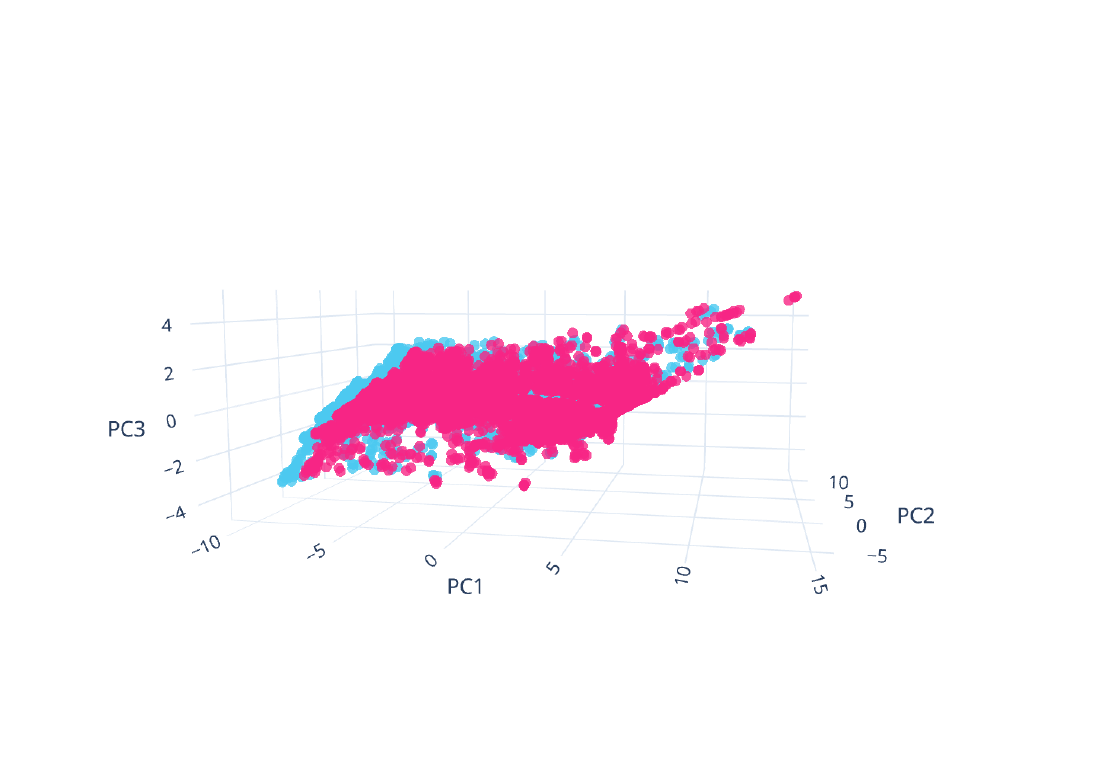}
        \caption{TTT- SSL decoder's penultimate emb.}
    \end{subfigure}
    \hfill
    \begin{subfigure}{0.48\textwidth}
        \centering
        \includegraphics[width=\linewidth]{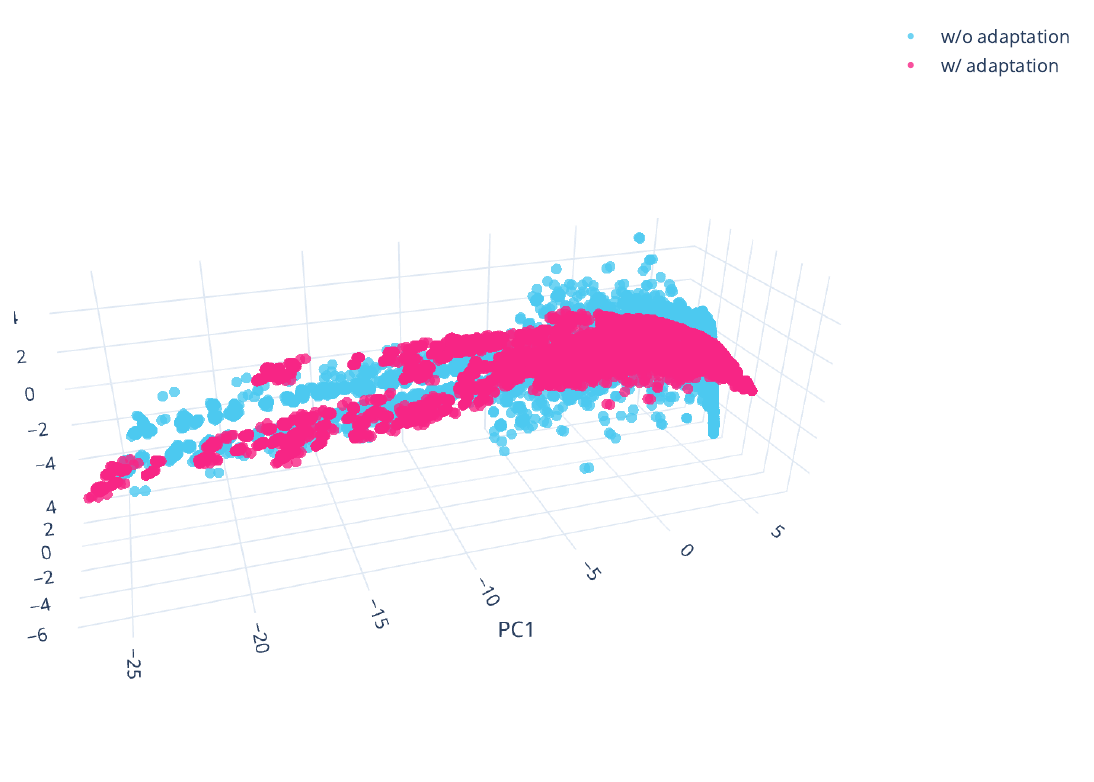}
        \caption{T3R- SSL decoder's penultimate emb.}
    \end{subfigure}

    \caption{\textbf{3D PCA visualization}. \textcolor[HTML]{4CC9F0}{Cyan} showcases reduced features without adaptation, while \textcolor[HTML]{F72585}{pink} highlights adapted ones. The viewpoint is adjusted for clearer visualization. The feature points appear approximately symmetric about a central axis. After adaptation, this symmetry becomes more apparent: the separation widens, and the tails grow sharper and more elongated. The exceptions are the SSL heads, whose shape remains nearly identical.
    }
    \label{fig:visualization}
\end{figure}

To observe the effects of test-time training, we visualize 3D PCA projections of embeddings at key positions, including the encoder's final layer and the second-to-last layers of both the main and \gls{ssl} decoders, as shown in Figure~\ref{fig:visualization}. 
Surprisingly, although \gls{ttt} does not employ a rotation matrix, its embedding space appears visually symmetrical. However, this symmetry does not expand further after adaptation, particularly in the encoder’s final layer. In contrast, \gls{t3r} amplifies the symmetric structure more intensively. This could be attributed to rotation matrices.
In terms of \gls{ssl} heads, the embedding space is unlikely to change under either approach, as the updates are still driven by the original \gls{ssl} loss at test-time.

\end{document}